%% file: main.tex
\documentclass[10pt,twocolumn,letterpaper]{article}
\usepackage[pagenumbers]{iccv} % To force page numbers, e.g. for an arXiv version

\usepackage{graphicx}
\usepackage{amsmath}
\usepackage{amssymb}
\usepackage{booktabs}
\usepackage{float}
\usepackage{colortbl}
\definecolor{Teal}{RGB}{80,139,243}
\definecolor{MyOr}{RGB}{250,190,88}
\definecolor{MyGr}{RGB}{77,175,124}
\definecolor{MyRed}{RGB}{226,106,106}
\definecolor{MyPur}{RGB}{148,124,176}
\usepackage[pagebackref,breaklinks,colorlinks,citecolor=Teal]{hyperref}

\usepackage[capitalize]{cleveref}
\crefname{section}{Sec.}{Secs.}
\Crefname{section}{Section}{Sections}
\Crefname{table}{Table}{Tables}
\crefname{table}{Tab.}{Tabs.}

\def\paperID{8159} % *** Enter the Paper ID here
\def\confName{ICCV}
\def\confYear{2025}

\title{Video Motion Graphs}

\author{Haiyang Liu\textsuperscript{1,2}\footnotemark[1]\quad 
Zhan Xu\textsuperscript{2}\quad 
Fa-Ting Hong\textsuperscript{2} \quad 
Hsin-Ping Huang\textsuperscript{2} \quad 
Yi Zhou\textsuperscript{2} \quad 
Yang Zhou\textsuperscript{2} \quad 
\\
\textsuperscript{1}The University of Tokyo \quad
\textsuperscript{2}Adobe Research  \quad
}

\begin{document}

\twocolumn[{%
\renewcommand\twocolumn[1][]{#1}%
\maketitle
\begin{center}
    \centering
    \captionsetup{type=figure}
    \vspace{-0.5cm}
    \includegraphics[trim=0 0 0 0, clip,width=1.0\textwidth]{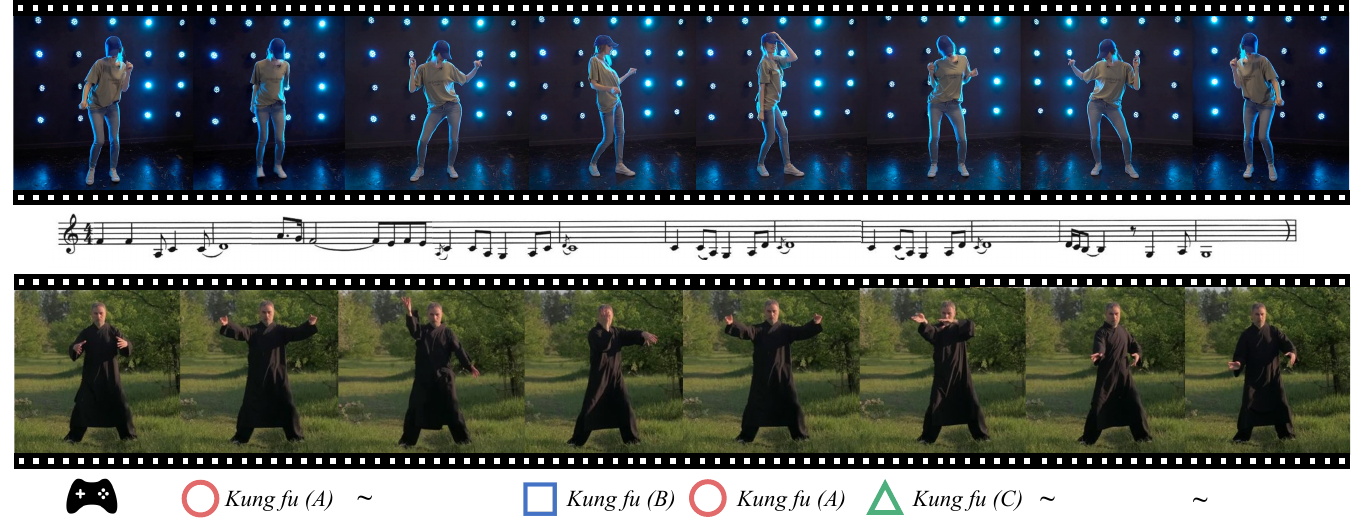}
    \captionof{figure}{\textbf{Video Motion Graphs} is a system to generate human motion videos from a reference video and conditional signals such as music, action tags and sparse keyframes. The video is generated by first retrieving matched video clips from reference video and then generating interpolation frames between clips to smooth the transition boundaries.}
    \label{fig:teaser}
    
\end{center}%
}]

\footnotetext[1]{Work done during Haiyang, Fa-Ting, and Hsin-Ping's internship at Adobe Research from 24/06 to 24/09.}

\begin{abstract}
We present \textit{Video Motion Graphs}, a system designed to generate realistic human motion videos. Using a reference video and conditional signals such as music or motion tags, the system synthesizes new videos by first retrieving video clips with gestures matching the conditions and then generating interpolation frames to seamlessly connect clip boundaries.
The core of our approach is \textit{HMInterp}, a robust Video Frame Interpolation (VFI) model that enables seamless interpolation of discontinuous frames, even for complex motion scenarios like dancing. HMInterp i) employs a dual-branch interpolation approach, combining a Motion Diffusion Model for human skeleton motion interpolation with a diffusion-based video frame interpolation model for final frame generation. ii) adopts condition progressive training to effectively leverage identity strong and weak conditions, such as images and pose. These designs ensure both high video texture quality and accurate motion trajectory.
Results show that our Video Motion Graphs outperforms existing generative- and retrieval-based methods for multi-modal conditioned human motion video generation. Project page can be found \href{https://h-liu1997.github.io/Video-Motion-Graphs/}{here}.
\end{abstract}

\section{Introduction}
Human motion videos play a vital role across numerous industries, including entertainment, virtual reality, and interactive media. However, capturing high-quality, realistic motion videos can be labour-intensive and costly. Recent advancements in video generation offer solutions in generating human motion videos based on inputs like skeletal animation, action labels, and speech audios, making production more efficient and customizable.

Human motion video generation has two primary approaches: generative- and retrieval-based methods. Generative-based models \citep{corona2024vlogger,wang2024dance,jiang2023text2performer,hu2023animate} synthesize all frame pixels from conditional inputs, offering flexibility in generating diverse motions. However, they often produce artifacts for complex contents, such as human distorted limbs, fingers, etc \citep{blattmann2023stable,yang2024cogvideox}. Retrieval-based models utilize key frames from given reference videos, and generate interpolation frames to ensure smooth transitions \citep{zhou2022audio,liu2024tango}. While they require reference material, they typically deliver higher video quality and maintain the actor's identity. Motivated by the video quality, we focus on the retrieval-based method for real-world applications. Existing retrieval-based methods, such as, Gesture Video Graph (GVR) \citep{zhou2022audio} and TANGO \citep{liu2024tango}, are specifically designed for co-speech gesture video generation and are not applicable to general human motion animation, \textit{e.g.}, dancing, kung fu, etc. In particular, these methods retrieve video frames based on input audio using a motion graph framework \citep{kovar2008motiongraphs}, and then apply a Video Frame Interpolation (VFI) model to generate the interpolated frames between the retrieved frames.
Extending these methods to a comprehensive system to accommodate diverse conditions beyond speech audio presents one main challenges: Their VFI model leverages the linear blended motion guidance and thus limits its ability to handle complex, dynamic motions like human dancing. In this paper, we proposed an improved diffusion-based VFI model, \textit{HMInterp}, to complete the essential functionality of the general video motion graphs system.
% Extending these methods to general human motions presents two main challenges: i) the lack of a comprehensive system to accommodate diverse conditions beyond speech audio, and ii) the VFI model leverages the linear blended motion guidance and thus limits its ability to handle complex, dynamic motions like human dancing. In this paper, we proposed two core contributions: an extended motion retrieval system via motion graph and an improved diffusion-based VFI model.

The motivation of \textit{HMInterp} is to seamlessly connect retrieved frames. Unlike the original VFI module in GVR, which uses a simple linearly interpolated motion as guidance in a flow-warp-based VFI model, HMInterp addresses the limitations of linear interpolation for complex, dynamic motions. For instance, approximately 78\% of mild speech gestures can be reasonably approximated through linear blending, while only 17\% of dance gestures which complex motion can, highlighting the need for a more advanced approach for human dynamic motion sequences (see supplemental for details).

Specifically, \textit{HMInterp} starts from UNet-based pretrained text-to-video model, AnimateDiff\cite{guo2023animatediff}, and further consists of a diffusion-based VFI model with motion guidance from a dedicated Motion Diffusion Model (MDM). HMInterp generates smooth interpolated frames while preserving accurate human structure and motion, guided by the MDM module. Based on \citep{shafir2023human}, the MDM is trained within a validated human skeleton space, ensuring both structural integrity and continuous motion of body parts. Moreover, we observed straightforward multi-condition (image and pose) joint training yields identity inconsistent results. To solve this, we introduce condition progressive training. It adopts different training orders and iterations for strong and weak conditions such as pose and image, respectively. Finally, to enhance the performance, we incorporate the ReferenceNet from \citep{hu2023animate,xu2024magicanimate} and reference decoder from \citep{xing2024tooncrafter}. As a consequence, these components enable \textit{HMInterp} to produce realistic interpolated frames. 

% we incorporate the ReferenceNet from \citep{hu2023animate,xu2024magicanimate} to maintain identity consistency and develop an enhanced reference decoder inspired by \citep{xing2024tooncrafter} to boost the interpolated frame quality. As a consequence, these components enable HMInterp to produce realistic interpolated frames, completing the essential functionality of the proposed Video Motion Graphs system.

The quality of \textit{HMInterp} enable us to implement the \textit{Video Motion Graphs} for general human video representation (in Figure \ref{fig:teaser}). From an engineering perspective, we complete the audio-only GVR baseline \cite{zhou2022audio} into a more comprehensive system. In particular, we define a standard four-stage pipeline for graph initialization, searching, frame interpolation and background reorganization. Besides, we adopt task-specific rule-based matching for searching, and introduce keyframe-based editing. This allows the system to retrieve relevant video clips to align with conditions like music beats. It also enables users to replace or customize specific frames by manually editing the output frames.

Overall, our contributions can be concluded as follows:

\begin{itemize}
    % \item We promote the \textit{Video Motion Graphs}-based retrieval system to general human motion videos. The system supports sequence and key-frame level control of reference videos with dynamic backgrounds. It shows diverse applications like real-time generation and key-frame editing, achieving state-of-the-art performance.  \zhan{since the title is Video Motion Graphs, let's not say Video Motion Graphs-based xxx}
    \item We propose \textit{Video Motion Graphs}, a comprehensive retrieval + generation system for general human motion videos. This system enables both sequence and keyframe retrieval, supporting applications such as real-time video generation and keyframe editing. It achieves state-of-the-art performance in generating high-quality, customizable human motion videos.
    \item We propose \textit{HMInterp}, a high-quality motion-aware, video frame interpolation module with the proposal that i) utilizes Motion Diffusion Model for generative and controllable human structure guidance. ii) adopts condition progressive training to effectively leverage identity strong and weak conditions. 
    % \item We introduce simple yet effective auxiliary designs, including a transformer-UNet architecture in the MDM and a reference-based decoder in HMInterp, to enhance both the generated motion and video frame quality.
\end{itemize}

\begin{figure*}[t]
\begin{center}
\includegraphics[trim=0 0 0 0, clip,width=1.0\textwidth]{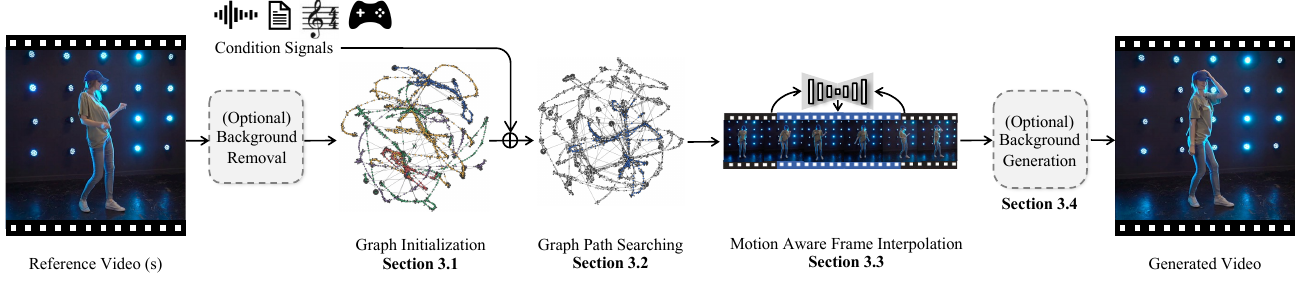}
\end{center}
\vspace{-0.3cm}
\caption{\textbf{System Pipeline of Video Motion Graphs.} Given input reference video(s) and a condition signal like music, Video Motion Graphs generates a new video in four steps: (i) representing the video as a directed graph, where nodes are RGB frames and edges indicate valid playback transitions, (ii) retrieving a frame playback path in the graph to match conditions based on task-specific rules, such as beat alignment, (iii) blending discontinuous frame transitions with a Motion-Aware Frame Interpolation Model, and (iv) optionally changing the video background through background removal and generation models.}
\label{fig:overview}
\end{figure*}

\section{Related Work}
\paragraph{Generative Human Motion Video Synthesis.} Generative methods generate all frames directly from networks with conditions. To the best of our knowledge, there is no unified model to accept flexible conditions and output general human motions.
% \zhan{our method also cannot support flexible condition, right?} 
There are task-specific models, such as generating motion video from text \citep{jiang2023text2performer}, music \citep{wang2024dance}, speech \cite{corona2024vlogger,lin2024cyberhost,he2024co,liu2022disco,liu2024emage,liu2022beat}, and pose \citep{zhang2023adding,hu2023animate,huang2024make,chan2019everybody,chang2023magicpose,zhu2024champ}. These methods are flexible to generate novel poses, but the video quality is sub-optimal. The video quality depends on the video generation backbone \citep{rombach2022high}. Even though the backbone has moved from UNet \citep{rombach2022high,chen2023videocrafter1,blattmann2023align,chen2024videocrafter2} to DiT \citep{esser2024scaling,hong2022cogvideo,yang2024cogvideox} the state-of-the-art video generation model such as Open-Source Stable Video Diffusion \citep{blattmann2023stable} still has broken hands and faces. This makes a video full of generated frames appear unnatural. Different from them, we only apply the generative model to a few frames and retrieve from reference video for the task to ensure higher video quality.

\paragraph{Retrieval Human Motion Video Synthesis.} GVR \citep{zhou2022audio} and TANGO \citep{liu2024tango} are previous works for Retrieval Motion Generation. They generate motion in three steps: (i) creating a motion graph based on 3D motion and 2D image domain distances, (ii) retrieval of the optimal path within this graph for the motion on the path is best matched to target speech, (iii) blending the discontinues frames by an interpolation network based on linear blended motion guidance.
However, their system is only designed for co-speech talkshow videos with stable and static backgrounds. We imporve the VFI to make it work on general human motions.

\paragraph{Video Frame Interpolation.} VFI is a classical low-level problem aimed at generating intermediate frames from beginning and end frames. The problem shifts from reconstruction to generative when motion dynamics increase. The fully end2end methods \citep{liu2017voxelflow, jiang2018superslomo, niklaus2018contextinterp, xue2019vetf, niklaus2020softmaxinterp, park2020bmbc, park2021asymmetricinterp, sim2021xvfi, wu2022optimizing, danier2022stmfnet, kong2022ifrnet, huang2022real, li2023amt, danier2024ldmvfi}, directly estimating middle frames using an end-to-end model trained on VFI tasks like VFI Diffusion \citep{jain2024video} or leverage pre-trained Text2Video models \citep{feng2025explorative}, such as DCInterp \citep{xing2025dynamicrafter}, which leverage the spatial and temporal patterns from VideoCrafter \citep{chen2023videocrafter1}. On the other hand, the solutions that rely on flow-warping or intermediate guidance show more promising results, FILM \citep{reda2022film} and VFIFormer \citep{lu2022video} are repetitive works with CNN and transformer backbones. However, as content differences increase, the estimated flow is not accurate enough and often leads to hands disappearing. To solve this, Pose-Aware Neural Blending \citep{zhou2022audio} and ACInterp \citep{liu2024tango} introduce explicit linear blended motion guidance with CNN and Diffusion. However the linear blending could not handle complex motions such as dance.  Unlike these, we leverage both VFI models and generative motion guidance to maintain video texture and motion correctness.

\section{Video Motion Graphs}
% As shown in Figure \ref{fig:overview}, \textit{Video Motion Graphs} represents video in a graph structure. Taking the input conditions and reference video, it generates video in four steps. It first represents reference video in a graph structure, where nodes are video frames and edges are valid transitions between frames (Section \ref{sec:graph_init}). Then, generating a video conditionally is converted to a graph path-searching problem (Section \ref{sec:graph_search}), where the optimal paths are generated video candidates. After searching, the Video Frame Interpolation (VFI) model is employed to smooth the discontinuous boundaries (Section \ref{sec:graph_vfi}). Optionally, a video background removal and video background generation model are employed to accept reference video with dynamic background \ref{sec:graph_background}).

The core idea behind the Video Motion Graphs is to represent an input video as a motion graph structure and synthesize output videos through motion graph search~\cite{zhou2022audio,liu2024tango}. Specifically, given a reference video and target conditional signals (\textit{e.g.}, speech audio, music, motion tags), the system generates video in four steps: (1) representing the reference video as a graph, with nodes as video frames and edges as valid transitions (Section \ref{sec:graph_init}); (2) framing conditional video generation as a graph path-searching problem, where path costs are guided by conditional signals (Section \ref{sec:graph_search}); (3) employing a Video Frame Interpolation (VFI) model to smooth discontinuous boundaries (Section \ref{sec:graph_vfi}); and (4) joining all searched and interpolated frames to construct the final output video. Optionally, background removal and generation models can be applied to handle reference videos with highly dynamic backgrounds (Section \ref{sec:graph_background}).

\subsection{Graph Initialization}
\label{sec:graph_init}
%We represent reference video as Graph \( G = \{ V, E \} \), where \( V \) contains meta-data such as RGB video frames, and \( E \) are valid transitions between vertices. 
We represent the reference video as a graph $\mathcal{G} = \{ \mathbf{V}, \mathbf{E} \}$, where each vertex $v\in\mathbf{V}$ denotes one video frame, and $e\in \mathbf{E}$ indicates if two frames (vertices) can be concatenated temporally with smooth transition. All original consecutive frames in the reference video are naturally connected by an edge. To measure the smoothness of the transition for other pairs of frames, we calculate the difference in human poses captured in frames. 
We followed GVR \citep{zhou2022audio} and TANGO \cite{liu2024tango} to compute the difference for 3D pose as \( d^{l}_{i,j} = || J^l_i - J^l_j ||_2 \).
Slightly different from TANGO and GVR, we replace 2D difference with \( d^{g}_{i,j} = || J^g_i - J^g_j ||_2 \), where \( J^i \) are 3D poses from any 3D pose detector. 
We then follow \cite{zhou2022audio} to pick a hyperparameter threshold $\tau$ based on the average nearest-neighbour distance calculated on the entire graph. To this end, an edge \( e=(v_i, v_j) \) exists if \( d^{l}_{i,j} + d^{g}_{i,j} \) is smaller than the threshold $\tau$. Finally, we follow the graph pruning in TANGO~\cite{liu2024tango} to remove dead-end nodes and obtain a final valid graph (an illustration can be found in Figure \ref{fig:overview}).

\begin{figure*}[t]
\begin{center}
\includegraphics[trim=0 0 0 0, clip,width=1.0\textwidth]{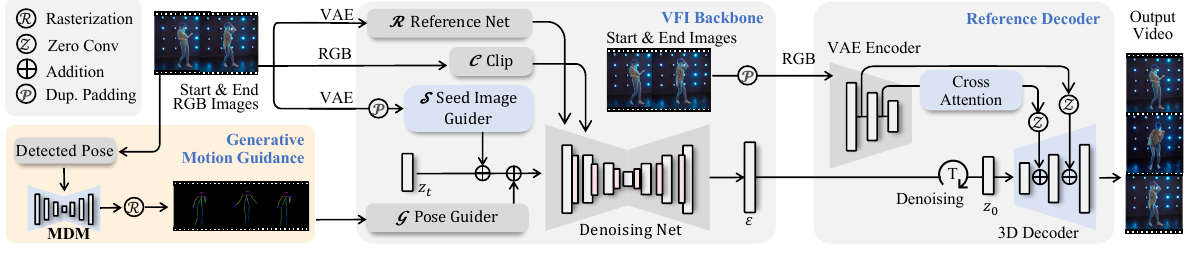}
\end{center}
\vspace{-0.5cm}
\caption{\textbf{HMInterp.} During inference, HMInterp takes start and end RGB images to generate interpolated video frames. It consists of three modules: (i) Generative Motion Guidance, which leverages our proposed MDM module to generate interpolated 2D poses and render them as RGB videos, (ii) the proposed VFI module that uses input images and poses to generate interpolation frames in latent space, and (iii) a Reference Decoder that combines denoised latent representations with input images to decode the final output video in pixel space through low-level feature injection.}
\vspace{-0.3cm}
\label{fig:hminterp}
\end{figure*}

\subsection{Path Searching}
\label{sec:graph_search}
% We formalize the video generation from $\mathcal{G}$ as a graph searching problem. Given a condition \( C \) such as speech audio, music or motion tag, we define sequence-level searching as minimizing the sum of the cost function. 
% % \( \min \sum \text{cost}(V_i, \omega) \). \zhan{$\omega$ is not defined, but i think this Eq is not necessary if not used in the following.} \yang{agree, let's remove it}
% The cost functions are task-specific scores for rule-based tasks, e.g., music beat matching score for music-to-dance video. We define 4 types of scores for action to motion, text to motion, music to dance, and speech to gesture. (See supplemental materials for details). The sequence-level searching could be computed by dynamic programming (DP) for offline and Beam Search real-time applications, respectively.

We formalize video generation from $\mathcal{G}$ as a graph search problem. Given a target signal, such as speech audio, music, or motion tags, we search for a valid graph path that minimizes the total path cost. This cost function combines (1) the intrinsic edge distance defined in Section~\ref{sec:graph_init} with (2) task-specific scores based on input conditions. In this paper, we define four video generation tasks suited to the proposed Video Motion Graphs: (1) human action generation via motion tags, (2) text-to-motion generation, (3) music-driven dance video generation, and (4) talking-avatar generation with gesture animation from speech audio. We have four dedicated heuristic-based searching solution for each of the four tasks (please check supp. for details.) It will define the per-frame path cost functions perform the sequence-level path searching with dynamic programming (DP) for offline processing or with efficient Beam Search~\cite{steinbiss1994improvements} for real-time applications. After searching, we can get a path with all necessary frames aligning well with the conditional signal, but it has discontinuous frames.

% We define the path cost functions as task-specific CLIP-like feature distance, \textit{e.g.}, CLIP-like joint embedding in MotionClip for text\cite{tevet2022motionclip}, ChoreoMaster \cite{chen2021choreomaster} for music, TANGO for speech audio \cite{liu2024tango}. Once the graph and path costs are defined, path searching can be performed using dynamic programming (DP) for offline processing or with efficient Beam Search~\cite{steinbiss1994improvements} for real-time applications. More details on task definitions and task-specific cost designs are provided in the supplementary materials. In addition, we introduce node-level path searching using shortest path algorithms on weighted graphs. Given the target sequence with \( K \) keyframes, we separately search the discontinuous region with the Dijkstra algorithm ~\cite{wang2011application}, using a length scale factor \( D \), which allows the target clip to be \( (1-D) \) times the length of the target, and then re-interpolate the searched path to the target length. 

% With both sequence-level and frame-level searching, our system can reassemble videos with various conditions.

\subsection{HMInterp for Video Frame Interpolation}
\label{sec:graph_vfi}
% In this section, we introduce HMInterp, which is the key module to make the Video Motion Graphs work. Our HMInterp, as shown in Figure \ref{fig:hminterp}, takes the input start and end frames and generates 12 intermediate frames (0.5 seconds at 24 FPS). Our key idea is to combine the Motion Diffusion Model (MDM) and Video Diffusion Model (VDM) \zhan{Video Diffusion Model seems to be a specific paper, which is not used in our method.} to fuse motion guidance from the MDM to VDM using classifier-free guidance, to maintain both motion correctness and video texture quality. We discuss the architecture design and training strategy below.
In this section, we introduce \textit{HMInterp}, the core enabling technology that allows Video Motion Graphs to seamlessly connect discontinuous frames, ensuring smooth motion transitions. As shown in Figure~\ref{fig:hminterp}, HMInterp takes start and end frames as input and generates 12 interpolated frames (equivalent to 0.5 seconds at 24 FPS). Unlike existing frame interpolation methods \cite{jain2024video,xing2025dynamicrafter,xing2024tooncrafter}, we propose a dual-branch interpolation approach that leverages both a Motion Diffusion Model (MDM)~\cite{shafir2023human} for human skeletal motion interpolation and a diffusion-based Video Frame Interpolation (VFI) for inbetweening frame generation. By fusing motion guidance from the MDM into the VFI with condition progressive training, HMInterp enhances both motion accuracy and high video texture quality. We first introduce the entire VFI Backbone for ensuring video textural quality, and then show how we generate and fuse the motion condition to allow correct motion trajectory.
\vspace{-0.4cm}

\begin{figure}
\begin{center}
\includegraphics[trim=40 0 40 0, clip,width=0.5\textwidth]{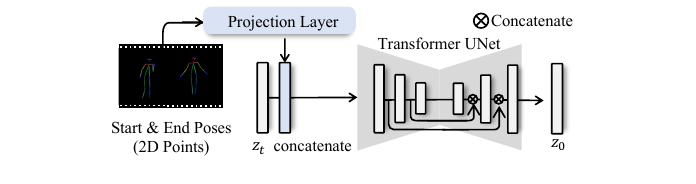}
\end{center}
\vspace{-0.5cm}
\caption{\textbf{Details of Motion Diffusion Model.} During training, MDM learns to reconstruct target interpolated 2D joint positions. It fuses the start and end poses through feature concatenation before the Denoising Transformer UNet. The vanilla Transformer in MDM is modified with skip connections and feature concatenation to enhance the details of the generated motion trajectory.}
\vspace{-0.5cm}
\label{fig:mdm}
\end{figure}

\paragraph{Video Frame Interpolation (VFI) Backbone.} We build our model on top of an existing UNet based text-to-video model~\citep{guo2023animatediff}, and adapt it for VFI task with additional conditioning signals, \textit{i.e.}, start and end frames. To enhance the performance, we also adopt the ReferenceNet from pose-to-video models\citep{hu2023animate,xu2024magicanimate}, replace the CLIP text encoder to CLIP image encoder\citep{radford2021learning},  and leverage the Reference Decoder from ToonCrafter\citep{xing2024tooncrafter}. Specifically, the start and end frames are encoded through a pre-trained VAE, duplicated padded to fill intermediate frames, and then fed as Seed Image Guiders in the VFI input (see Figure~\ref{fig:hminterp}). In addition to CLIP features, we further inject two types of low-level features to improve frame quality. Inspired from \citep{hu2023animate,xu2024magicanimate}, we use ReferenceNet to inject hierarchical latent feature guidance from the start and end frames, significantly enhancing identity and appearance fidelity. Additionally, for tasks involving full-body video generation, human details like faces may degrade at lower resolutions (\textit{e.g.}, 256×256) due to VAE decoder limitations. To mitigate this, we implement an improved Reference Decoder for realistic videos, which is originally proposed by ToonCrafter \citep{xing2024tooncrafter} for animation videos. It applies skip-connections from the VAE encoder low-level latents to boost detail retention. Unlike \cite{xing2024tooncrafter}, our reference decoder is initialized from the temporal decoder in Stable Video Diffusion \citep{blattmann2023stable}. In addition, we propose to use duplicated reference frames as input, which is simple but brings clear performance gain without extra additional inference costs. We compare the effectiveness of our re-implementation against \cite{xing2024tooncrafter} in Figure~\ref{fig:ablition_ref}. The above design allows the backbone to generate videos with high textural quality.

% Lastly, we condition the frame interpolation with 2D pose guidance. As shown in Figure \ref{fig:mdm}, during training, we train the VDM by the ground truth 2D pose guidance. During inference, we take the generated 2D pose from 2D MDM. Our key motivation to train the MDM on 2D points space is to generate trajectory-consistent motion from a loss perspective. i.e., for 2D position loss, point disappear will have a higher loss than bias from the correct trajectory. For the implementation of MDM, we use a Vanilla MDM to predict intermediate $\hat{X}^{t}$ from $X^{start}$ and $X^{end}$. Different from MDM, we noticed the motion details will be lost when using the vanilla 8-layer transformer in MDM. To solve this, we implement a UNet-like transformer architecture as shown in Figure \ref{fig:mdm}. It fuses the feature from shallow layer to deeper layer via concatenation and skip-connection. With the transformer-UNet, our MDM generates more correct non-linear motion interpolation trajectory.

\begin{table}
\centering
\caption{\textbf{Comparison of Human Motion Video Generation.}}
\resizebox{0.45\textwidth}{!}{%
\begin{tabular}{lcccc}
                          & PSNR $\uparrow$ & LSIPS $\downarrow$ & MOVIE $\downarrow$ & FVD $\downarrow$ \\ 
\hline
AnimateAnyone \citep{hu2023animate}            & 35.55     & 0.044     & 54.68     & 1.369      \\
MagicPose \citep{chang2023magicpose}               & 35.64     & 0.048      & 51.97      & 1.277      \\
UniAnimate \citep{wang2024unianimate}               & 36.75     & 0.042      & 49.89      & 1.090      \\
MimicMotion \citep{zhang2024mimicmotion}             & 36.30     & 0.047      & 46.84      & 1.078      \\
\hline
Ours ($f = 32$)      & \textbf{42.91}     & \textbf{0.009}      & \textbf{37.31}      & \textbf{0.180}      \\
Ours ($f = 64$)      & 42.75     & 0.010      & 37.53      & 0.213      \\
Ours ($f = 216$)      & 39.75     & 0.029      & 39.89      &  0.799     \\
\end{tabular}}
\label{tab:videopsnr}
\end{table}

\paragraph{Motion Diffusion Model (MDM).} We then introduce the MDM that generates interpolated 2D poses between start and end frames. These interpolated 2D poses serve as the final conditioning input for the VFI, enabling it to produce more accurate human poses with valid structures. The existing MDM ~\cite{shafir2023human} often loses motion details when using a vanilla 8-layer transformer. To address this, we implemented a UNet-like transformer architecture (see Figure \ref{fig:mdm}), which fuses features from shallow to deeper layers using concatenation and skip-connection. This design allows our MDM to generate more accurate, non-linear motion interpolation trajectories (see ablation in Table \ref{tab3:ab_mdm}). During training, we guide the VDM with ground truth 2D poses, while during inference, we condition it on generated 2D poses from the MDM.

\paragraph{Condition Progressive training.} We first train the Reference Decoder and MDM separately and then freeze the VAE, MDM and CLIP. The remaining trainable parameters are ReferenceNet, Seed Image Guider, Pose Guider, and Denoising Net. For Denoising Net we load the pretrained weights from AnimateDiff. The straightforward implementation is training the image conditions (ReferenceNet, Seed Image Guider) together with pose conditions (Pose Guider). However, this yield facial appearance inconsistent between the generated and groundtruth frames (See Figure \ref{fig:combined} for details). To solve this, we consider training different conditions progressively: We first conduct Seed Pre-Training, train the VFI module with image condition only for long iterations (100k), to ensure the interpolated frames follow the reference appearance accurately. then, we apply Few-Step Pose Finetuning, combine the image and pose conditions to train VFI module with full conditions for a few iterations (8k). In our experiments, we observed that the other implementations, including swapping the training order, fine-tuning with pose condition only, or fine-tuning with pose guidance for more steps will impact human identity preservation. The condition progressive training could mitigate this effect.

% The reason for separating and using fewer steps to train the pose video is due to the identity lack of the pose module, i.e., human’s identity will change slightly when training with pose guidance for longer durations. 
% Our assumption is that the identity lock model maintains the human appearance for the current pose, attributed to the pose module, and we reduce the effect of the pose module by separated training and fewer steps in the pose video stage. 
\paragraph{Loss Functions.} VFI and MDM modules are trained with $v$-predication and $x_{0}$-prediction respectively. The Reference Decoder is trained with MSE and perceptual loss. 
% \zhan{can we say step 1,2,3?}

\begin{table}
\centering
\caption{\textbf{User Study Win Rate.} Subjective comparison between our model and baselines across different motion generation tasks.}
\resizebox{0.45\textwidth}{!}{%
\begin{tabular}{lccc}
Preference of Ours                        & Dance \citep{wang2024dance} & Gesture \citep{he2024cospeech} & Action \citep{jiang2023text2performer} \\ 
\hline
Texture Quality   & 82.10\%              & 78.38\%             & 69.12\%          \\
Cross-Modal Align.& 88.39\%          & 47.63\%             & 45.21\%          \\
Overall Preference & 84.99\%      & 70.24\%             & 61.05\%          \\
\end{tabular}}
\label{tab:subjective_evaluation}
\end{table}

\begin{table}
\centering
\caption{\textbf{Reference Video Length Impact.} Ablation study with FVD, Motion Diversity, and Frame Consistency (LPIPS).}
\vspace{-0.3cm}
\resizebox{0.95\linewidth}{!}{
\begin{tabular}{lccc}
          & FVD $\downarrow$ & Motion Div. $\uparrow$ & FC (LPIPS) $\downarrow$ \\ \hline
DanceAnyBeat \citep{wang2024dance}              & 1.981                     & 3.669                                  & 0.0993                                         \\
Ours (10s database)       & 0.611                     & 2.636                                  & 0.0466                                         \\
Ours (100s database)      & 0.497                     & 5.724                                  & 0.0418                                         \\
Ours (1000s database)     & \textbf{0.413}                     & \textbf{6.024}                                  & \textbf{0.0427}                                         \\ \hline
Real Video                & -                         & 5.951                                  & 0.0408                                         \\ 
\end{tabular}}
\label{tab:video_length}
\vspace{-0.4cm}
\end{table}

\subsection{Video Background Reorganization}
\label{sec:graph_background}
% With the path search and HMinterp, our Video Motion Graphs could generate realistic video with a nearly stable background. However, if the background is dynamic (e.g., running on the street), the background blending brings unrealistic artifacts to the final video. To solve this, we propose to use Video Background Removal and Video Background Generation \citep{pan2024actanywhere} for dynamic video backgrounds. The Video Motion Graphs will remove the background and generate foreground-only Motion Videos, and then generate a natural background with the background reference. Thanks to the powerful background module, we successfully disentangle the foreground and background, significantly increasing the application range of our Video Motion Graphs.
Our \textit{Video Motion Graphs} can generate realistic videos with the modules described above when reference videos have static or nearly stable backgrounds. However, when backgrounds are dynamic (\textit{e.g.}, running scenes on a street), the VFI module struggles to blend such diverse backgrounds, leading to unrealistic artifacts in the final video. To address this, we employ Video Background Removal \citep{kamel2008moving} to remove video dynamic background, then apply the proposed Video Motion Graphs to synthesis human foreground videos, and finally apply Video Background Generation \citep{pan2024actanywhere} to put back background or generate interesting novel backgrounds.

\begin{figure*}
\begin{center}
\includegraphics[trim=0 0 0 0, clip,width=1.0\textwidth]{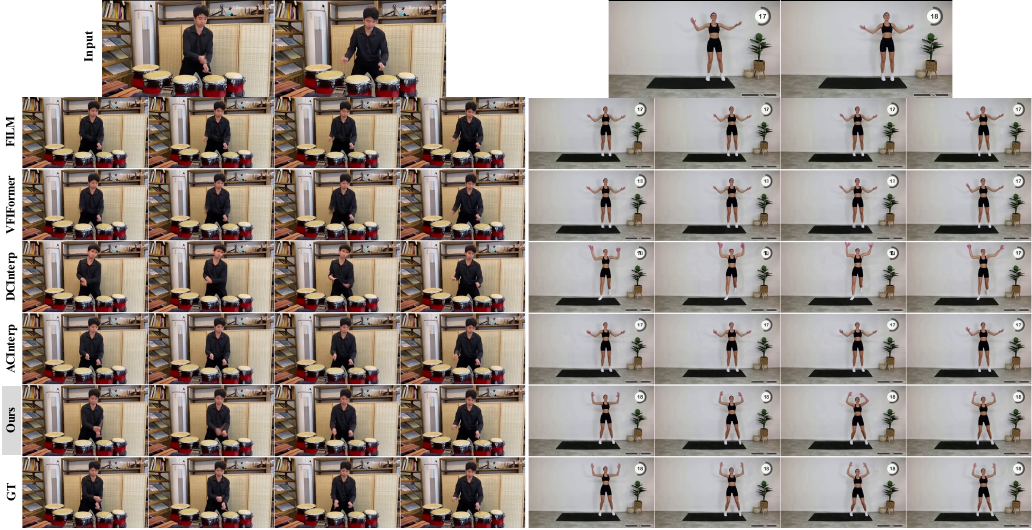}
\end{center}
\caption{\textbf{Subjective Results of HMInterp.} Compared to previous methods, HMInterp generates intermediate frames with accurate motion trajectories, addressing challenges that previous methods with linear motion guidance could not resolve. Top: dynamic motion, such as drumming. Bottom: self-loop motion for fitness activities.}
\label{fig:subjective_hminterp}
\end{figure*}

\section{Evaluation}
As our system is entirely novel, with no existing open-source methods available for direct comparison on this multi-modal conditioned human video generation task, we compare our method with separately baselines for different sub-tasks, besides, we provide qualitative end-to-end results and showcase various applications in the supplementary material.

\paragraph{Datasets.} We combine a series of video datasets focused on human motion to conduct the experiments. The datasets include MotionX \citep{lin2024motion} for general motion, Show-Oliver \citep{talkshow:yi2022generating} TED-Talk for talkshow, and Champ \citep{zhu2024champ} for dance videos. The datasets are around 100 hours for training, we filtered the data based on motion quality (see supplemental for details) for training. The evaluation is conducted on a randomly sampled 367 video test set.

% \textbf{Evaluation tasks.} Since there is no related unified model to handle different conditions and generate motion videos, we compare our method with separated state-of-the-art models for action, dance, talkshow, instruments, and sports videos. Then, we discuss the performance of HMInterp module by comparing it to previous video frame interpolation models. We further demonstrate the effectiveness of our improvement on MDM and Reference Decoder for HMInterp. Finally, we discuss the applications such as key-frame-based editing, real-time video generation based on the extended Video Motion Graphs system.

\subsection{Evaluation of Video Motion Graphs System}

We evaluate video generation quality on human motion videos. The generation quality is measured by objective video quality scores and subjective user studies. The video quality evaluation is shown in Table \ref{tab:videopsnr}. We compared our method with skeleton-pose driven video generation models, including \textit{Animate Anyone} \citep{hu2023animate}, \textit{Magic Pose} \citep{chang2023magicpose}, \textit{UniAnimate} \citep{wang2024unianimate}, and \textit{Mimic Motion} \citep{zhang2024mimicmotion}. We select the video length larger than 300 frames in the test set for evaluation. We use the starting frame as the reference frame and GT skeleton poses as input for those methods.
% For pose2video models, it conducts self-reconstruction from its 2d pose. 
For our Video Motion Graphs, we randomly mask out $f$ frames from input videos and recover them by HMInterp interpolation. 
% For example, for a 300-frame video with a drop frames $f = 32$, we random mask 4 x 8 continues segments and replace them by interpolation results of HMInterp. 
We evaluated the results when masking out $f=32, 64, 216$ frames, where higher values indicate a greater challenge for interpolation. 
The result in Table \ref{tab:videopsnr} shows that 
% the generally sampled video from our system, i.e., with $f$ around 32 for a $10\%$ interpolation, 
ours ($f=32$) significantly outperforms the generated models in all objective terms. Besides, even when it comes to the hardest case with $f=216$, our method still outperforms existing pose2video models.

We also evaluated the motion alignment and overall performance of generated videos via a subjective user study. We compared different baselines given the specific task. We compared with DanceAnyBeat \citep{wang2024dance} for music-driven human dance generation, S2G-Diffusion \citep{he2024cospeech} for audio-driven human speech gesture animation, and Text2Performer \citep{jiang2023text2performer} for action2video (details are in supplement material). We generated 240 videos, 80 per task, as the results and conducted theuser study via Google Forms. For each task, users evaluate 10 videos randomly sampled for each task, 40 videos for all tasks. The results are averaged for all 82 users and shown in Table \ref{tab:subjective_evaluation}. From the table, we observe that the Video Motion Graphs achieves comparable performance in alignment and gains higher user preference due to notable improvements in video texture quality.

One common limitation for all retrieval-based methods that the target should be aligned with the database, such as image retrieval. But in our case, as shown in Table \ref{tab:video_length}, it will outperform current generative models when the database is larger than 100s, which is relatively easy to record. We report the objective scores below via FVD, Motion Div. (2d joints positions diversity), and Frame Consistency (LPIPS frame difference).

\begin{table}
\centering
\caption{\textbf{Objective comparison of VFI methods.} $s$ denotes the number of start and end frames. Our HMInterp with $s = 1$ outperforms previous non-diffusion-based methods, FILM and VFIFormer, as well as diffusion-based methods, DCInterp and ACInterp. Additionally, HMInterp with $s = 1$, our main model integrated into the Video Motion Graphs, demonstrates high scores at both pixel and feature levels.}
\resizebox{0.45\textwidth}{!}{%
\begin{tabular}{lcccc}
                          & PSNR $\uparrow$ & LPIPS $\downarrow$ & MOVIE $\downarrow$ & FVD $\downarrow$ \\ 
\hline
FILM \citep{reda2022film} & \underline{37.57} & 0.043 & \underline{40.64} & 1.303 \\
VFIFormer \citep{lu2022video} & 36.29 & 0.057 & 52.32 & 1.390 \\
DCInterp \citep{hu2023animate} & 36.73 & 0.051 & 48.62 & 1.374 \\
ACInterp \citep{zhou2022audio} & 37.57 & \underline{0.040} & 46.94 & \underline{1.280} \\
\hline
HMInterp ($s = 1$)  & \textbf{39.53} & \textbf{0.034} & \textbf{39.18} & \textbf{1.210}  \\ 
HMInterp ($s = 3$)  & \textbf{40.40} & \textbf{0.021} & \textbf{37.85} & \textbf{0.619}  \\
\end{tabular}
\vspace{-0.6cm}
}
\label{tab3:interp}
\end{table}

\begin{figure*}
\begin{center}
\includegraphics[trim=0 0 0 0, clip,width=1.0\textwidth]{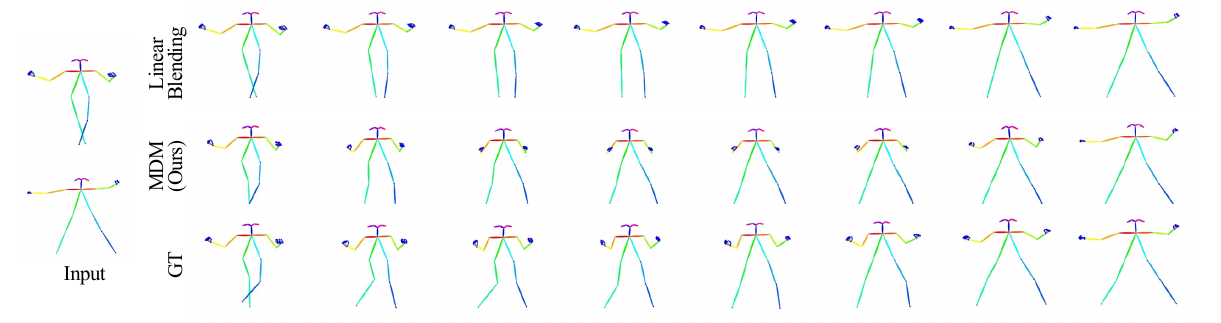}
\end{center}
\vspace{-0.5cm}
\caption{\textbf{Ablation of Generative Motion Blending.} Compared to linear blending, our Motion Diffusion Model-based blending generates non-linear intermediate motion for highly dynamic actions, such as dance. Top: linear blending consistently raises the hands above the shoulders. Middle: our blending presents a more complex and natural interpolated dance motion.} 
\label{fig:subjective_mdm}
\end{figure*}

\subsection{Evaluation of HMInterp}
We compare the video frame interpolation performance of HMInterp with previous non-diffusion-based frame interpolation methods FILM \cite{reda2022film}, and VFIFormer \citep{lu2022video}, and diffusion-based methods DCInterp \cite{xing2025dynamicrafter} and ACnterp \cite{liu2024tango}. The objective metrics include MOVIE \cite{seshadrinathan2009motion}, FVD \cite{skerry2018towards} for video quality, and IS, PSNR for image quality (see supplemental for metric details). The results are shown in Table \ref{tab3:interp}. Our HMInterp outperforms baselines on all terms. Some qualitative comparisons are shown in Figure \ref{fig:hminterp}. We highlight a limitation of methods like FILM and VFIFormer, which tend to miss generating body regions due to the absence of explicit guidance.. Other methods, such as ACInterp with linear interpolation, often produce incorrect motion trajectories. In contrast, our approach maintains both high video quality and accurate motion trajectories.

\begin{table}
\centering
\vspace{-0.3cm}
\caption{\textbf{Ablation Study of HMInterp.} Without motion guidance, performance shows a clear drop in feature-level metrics, such as ISIPS and FVD. Without the low-level reference decoder, pixel-level metrics, such as PSNR and MOVIE, decrease.}
\resizebox{0.45\textwidth}{!}{%
\begin{tabular}{lcccc}
                          & PSNR $\uparrow$ & LPIPS $\downarrow$ & MOVIE $\downarrow$ & FVD $\downarrow$ \\ 
\hline
HMInterp ($s = 1$)  & \textbf{39.53} & \textbf{0.034} & \textbf{39.18} & \textbf{1.210} \\
w/o motion guidance  & 39.17 & 0.048 & 41.34 & 1.391 \\
w linear guidance  & 39.16 & 0.042 & 41.06 & 1.297 \\
w/o reference decoder  & 37.21 & 0.039 & 49.67 & 1.283 \\
w zero reference decoder  & 38.13 & 0.034 & 40.11 & 1.221 \\
\end{tabular}}
\vspace{-0.5cm}
\label{tab3:ab_mdm}
\end{table}

\begin{figure}
\begin{center}
\includegraphics[trim=0 0 0 0, clip,width=0.48\textwidth]{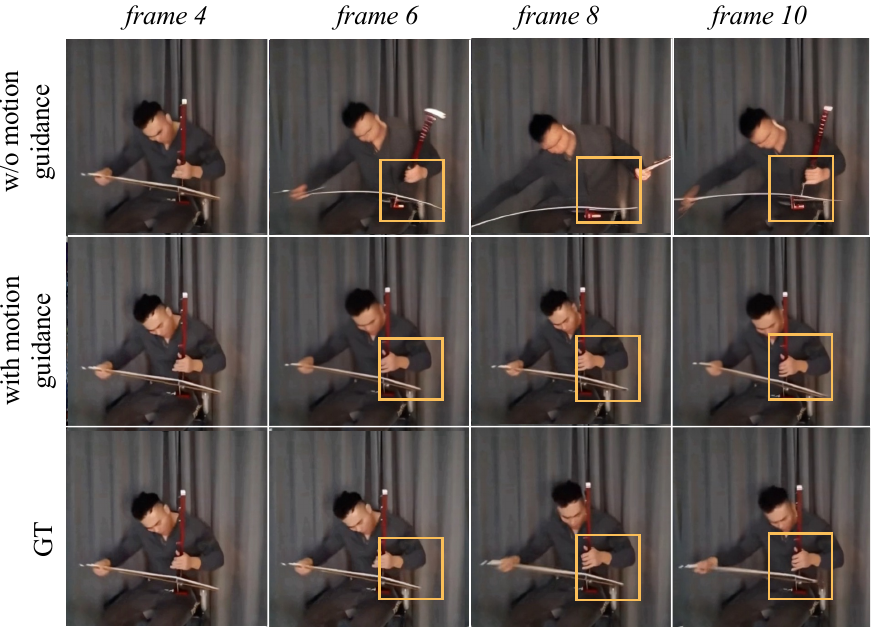}
\end{center}
\vspace{-0.3cm}
\caption{\textbf{Additional benefits of motion guidance.} Motion guidance distinguishes regions to obtain the correct interaction with the image content. Top: without motion guidance, BanHu's texture was broken due to out-of-range motion. Middle: with motion guidance, both human motion and BanHu display the correct texture.}
\vspace{-0.5cm}
\label{fig:ablition_mdm}
\end{figure}

\vspace{-0.3cm}
\paragraph{Ablation of Motion Diffusion Model.} 
As shown in Table \ref{tab3:ab_mdm} and Figure \ref{fig:ablition_mdm}, we show the difference that without the MDM, the results degrade on all terms, and the model subjective tends to refer to features from incorrect regions. On the other hand, we demonstrate our MDM is better than linear blending in Figure \ref{fig:mdm}.

\begin{figure*}
\begin{center}
\includegraphics[trim=0 0 0 0, clip,width=1.0\textwidth]{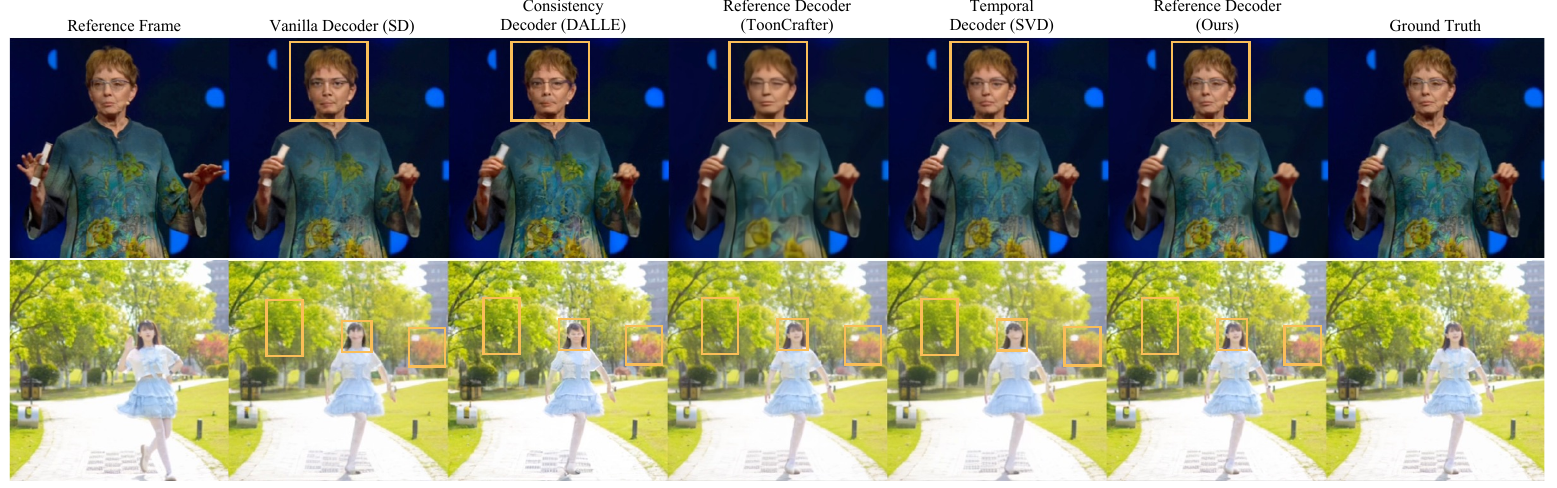}
\end{center}
\vspace{-0.3cm}
\caption{\textbf{Ablation of Reference Decoder.} We adopt duplicated padding inputs based on the reference decoder in ToonCrafter. It significantly improves results at lower resolutions, \textit{e.g.}, 256 $\times$ 256. Compared to other methods, our implementation provides accurate facial (top) and background (bottom) details.}
\label{fig:ablition_ref}
\end{figure*}

% \begin{figure}
% \begin{center}
% \includegraphics[trim=-5 0 0 0, clip,width=0.5\textwidth]{fig/fig11_v1_1.pdf}
% \end{center}
% \vspace{-0.3cm}
% \caption{\textbf{Background Reorganization.} Combined with video background removal and generation methods, our Video Motion Graphs system could disentangle the foreground and background. }
% \vspace{-0.4cm}
% \label{fig:background}
% \end{figure}
%\vspace{-0.3cm}

% \paragraph{Effectiveness of Condition Progressive Training.} 
% We compare our condition progressive training with variance in Table \ref{tab:condtion_progressive} and Figure \ref{fig:combined}. The straightforward pose-to-video training \cite{hu2023animate,xu2024magicanimate}, \textit{i.e.}, using pose as stage 1, could generate plausible results, but the appearance still remains inconsistency. Addressing this inconsistency is critical for production-level, artifact-free results. The other variances, \textit{e.g.}, row 2-5 in Table \ref{tab:condtion_progressive}, demonstrate that the pose condition will decrease the consistency. Our Condition Progressive Training that leveraging identity strong condition (seed images) first and identity weak condition (pose) later shows best performance.   

\begin{figure}
  \centering
  \includegraphics[width=\columnwidth]{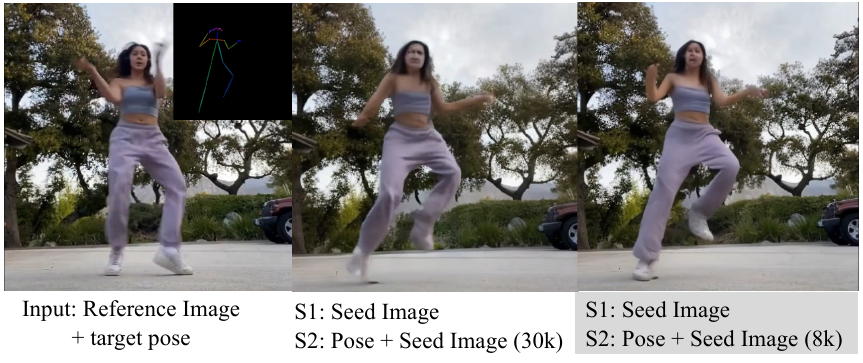}
  \caption{\textbf{Subjective comparison for progressive training}. The S1 and S2 denote training stages. After comparing different condition selections for each stage, we observed that training with pose conditions more will decrease the appearance consistency. Our progressive training (in gray) shows highly consistent results with the groundtruth. Refer to Table \ref{tab:condtion_progressive} for objective scores.}
  \label{fig:combined}
\end{figure}

\paragraph{Effectiveness of Condition Progressive Training.}
We compare our condition progressive training with different variants in Table \ref{tab:condtion_progressive} and Figure \ref{fig:combined}. The straightforward pose-to-video training \cite{hu2023animate,xu2024magicanimate}, \textit{i.e.}, directly using the pose condition as the initial stage, generates plausible motion results but leads to notable appearance inconsistencies across frames. Addressing these inconsistencies is essential for achieving artifact-free and production-level video results. Other variants in Table \ref{tab:condtion_progressive}, such as sequential training with pose followed by seed image ($P\rightarrow SI$) and simultaneous training ($P+SI$), demonstrate that introducing pose conditions early or simultaneously with identity conditions (seed images) tends to degrade the consistency of generated videos. Specifically, simultaneous condition training ($P+SI$) slightly improves metrics over the pose-only baseline but still falls short compared to a progressive strategy. Our proposed Condition Progressive Training approach, where we first employ identity-strong conditions (seed images) and later incorporate identity-weak conditions (pose), achieves the best quantitative performance.
% Moreover, the performance peak at 8k iterations indicates an optimal training duration that effectively balances the two conditions. Training longer (30k iterations) slightly decreases performance. Visual comparisons in Figure \ref{fig:combined} further show that our method maintains identity consistency and visual coherence.

\begin{table}
\centering
\caption{\textbf{Ablation study for condition progressive training.} $P$ denotes pose condition (pose-to-video training in AnymateAnyone), $SI$ is Seed Image condition (start and end reference images), \textit{iters} is training iterations. Ours (in gray) progressive training shows best performance. See Figure \ref{fig:combined} for subjective results.}
\resizebox{0.95\linewidth}{!}{
\begin{tabular}{llcccc}
\textbf{Stage 1}  & \textbf{Stage 2} & \textbf{PSNR $\uparrow$} & \textbf{LPIPS $\downarrow$} & \textbf{MOVIE $\downarrow$} & \textbf{FVD $\downarrow$} \\ \hline
$P$ & -                     & 35.55  & 0.044  & 54.68 & 1.369 \\
$P$ & $SI$                  & 36.81  & 0.041  & 51.66 & 1.330 \\
$P$ + $SI$ & -              & 36.62  & 0.041  & 52.41 & 1.325 \\
$SI$ & -                    & 36.84  & 0.043  & 51.89 & 1.339 \\ 
$SI$ & $P$                  & 36.83  & 0.042  & 52.03 & 1.336 \\ 
\rowcolor{gray!15}
$SI$ & $P$ + $SI$ (8k \textit{iters}) & \textbf{37.21} & \textbf{0.039} & \textbf{49.67} & \textbf{1.283} \\
$SI$ & $P$ + $SI$ (30k \textit{iters}) & 36.87 & 0.041 & 51.69 & 1.307 \\ 
\label{tab:condtion_progressive}
\end{tabular}}
\vspace{-0.3cm}
\end{table}

\paragraph{Effectiveness of The Improved Reference Decoder.} 
As shown in Table \ref{tab3:ab_mdm}, the reference decoder brings low-level consistency. Due to the memory issue, we keep the single-frame input on vanilla ReferenceNet, but we propose to use duplicated frames instead of zero padding frames on Reference Decoder. Compared with the baseline ToonCrafter \citep{xing2024tooncrafter}, this brings a clear benefit with a PSNR improvement over 1.0. The subjective results are shown in Figure \ref{fig:ablition_ref}, compared with baseline decoders. Such as vanilla decoder (Stable Diffusion) \citep{rombach2022high}, Consistency Decoder (DALL-E) \cite{ramesh2021zero}, Temporal Decoder (Stable Video Diffusion) \citep{blattmann2023stable}, and DualReference Decoder (ToonCrafter) \citep{xing2024tooncrafter}, our Reference Decoder could decode a background consistent video.

\subsection{Applications of the System}
\paragraph{Real-time video generation.} By pre-caching the possible transition results in the memory, our method could generate video in real-time without length limitation. It's recommended to see the real-time kungfu demo to demonstrate the potential applications like video game apps for our method.

\paragraph{Key-frame editing.} Our system supports generating the target path based on keyframes, (see supplement materials for details). The generated video could remain aligned with the predefined keyframes.
By replacing the motion guidance generated by MDM with a custom 2D pose video, our HMInterp could generate different motions with the reference video. Make it possible to combine the generative model with the retrieval model for further applications.

% \textbf{Video Style Mix for Different Speakers.} If the Video Motion Graphs is created with video from different speakers, it could generate some funny applications, such as switching speakers within the same automatic scene with the same actions or matching to the music beat.
% \vspace{-0.3cm}

% \paragraph{Video background removal and synthesis.} Replace the video background with background removal and generation methods~\cite{pan2024actanywhere} unlock the power of Video Motion Graphs to support dynamic background videos, such as the example shown in Figure \ref{fig:background}.
% Our Video Motion Graphs could focus on foreground only and post-process the background to handle the dynamic background.

% \subsection{Limitation}
% Different from full-generative methods that only require a single reference image, our system requires reference video to generate the video quality. The video’s motion diversity is highly reliant on the length of reference video. We found 1-min reference video is typically required to generate a natural video. In the future, combining generative methods to generate fake frames with automatic quality filtering are potential to solve this problem.
% Besides, the current method adopts task-specific rules for video retrieval. A unified condition representation model based on LLM may make broader applications for this system.

\section{Conclusion}

In this paper, we extend the previous speech gesture Video Motion Graphs system to a more robust and versatile framework for general human motion videos. By introducing HMInterp, a significantly enhanced video frame interpolation model, we enable seamless blending of open-domain human motion videos. It supports a range of applications including keyframe editing, and real-time video generation.

{
    \small
    \bibliographystyle{ieeenat_fullname}
    \bibliography{main}
}

\input{sup}

\end{document}

%% file: sup.tex
% \appendix

% --- PDF will be split by an editor (e.g. macOS preview), so need to restart from page 1
\setcounter{page}{1}

% --- repeat the title (AT: haven't found a more elegant way to do this...)
\twocolumn[
\centering
\Large
\textbf{Video Motion Graphs} \\
\vspace{0.5em}Supplementary Material \\
\vspace{1.0em}]

 %< twocolumn
\appendix
% \paragraph{Appendix}
\noindent This supplemental document contains five sections: 
\begin{itemize}[leftmargin=*]

\item Video and Codes (Section \ref{secsup5}). 

\item Importance of Non-linear Blending (Section \ref{secsup1}).

\item Implementation Details (Section \ref{secsup6}).

\item Rule-based Searching (Section \ref{secsup4}).

\item Baseline Settings (Section \ref{secsup3}).

\item Evaluation Metrics (Section \ref{secsup2}).

\item Ablation study of CLIP and Reference Net (Section \ref{secsup7}).

\end{itemize}

\section{Video and Codes}
\label{secsup5}

We provide a comprehensive video and separate videos to demonstrate the performance of our system, including:

\begin{itemize}
    \item General results for dance, gesture, and action-to-motion generation.
    \item Additional applications, including real-time multiple motion generation, keyframe editing and replacement.
    \item Comparison of different Video Frame Interpolation (VFI) methods.
    \item Evaluation of the effectiveness of the Motion Diffusion Model (MDM).
    \item Evaluation of the effectiveness of the Reference Decoder.  
\end{itemize}

The anonymous scripts including the Video Motion Graph system, the system first uses PoseInterp for motion interpolation and then employs HMInterp for video frame interpolation.

\section{Importance of Non-linear Blending}
\label{secsup1}

Linear blending in prior works such as GVR \cite{zhou2022audio} and TANGO \cite{liu2024tango}, offers a straightforward approach to guiding motion interpolation. While effective for simple motion scenarios like talk shows, it struggles with complex and dynamic motions, such as dance. This limitation underscores the need for non-linear blending techniques to achieve realistic motion generation. To illustrate this, we analyze the differences between linearly blended motion and ground-truth motion using sequences from \textit{Show-Oliver} \cite{talkshow:yi2022generating} and \textit{Champ-dance} \cite{zhu2024champ}. By setting a fixed threshold (e.g., 0.001) to measure deviations, we observe a clear discrepancy between linear blending and the ground truth, particularly for complex motions. Specifically, 78\% of the samples from \textit{Show-Olive} falls below the threshold, indicating that linear blending suffices for most cases in this dataset. However, only 17\%  of the samples from \textit{Champ-dance} is in the same threshold, highlighting that linear blending is unsuitable for high-dynamics motions.

\section{Implementation Details}
\label{secsup6}
We initialize our VFI model using the pretrained weights from Stable Diffusion 1.5 \cite{rombach2022high} for spatial layers and AnimateDiff \cite{guo2023animatediff} for temporal layers. All the training begins with a learning rate of \(1 \times 10^{-5}\).  The VFI model (both spatial and temporal layers) is initially trained without motion guidance at a resolution of \(256 \times 256\) on Nvidia H100 GPUs. This training is conducted for 100K iterations. Next, we train the VFI model with pose guidance, keeping the weights of the image guider fixed, at a resolution of \(512 \times 512\) for 8K iterations. The reference decoder is trained separately for 100K iterations. The Motion Diffusion Model (MDM) is trained on Nvidia A100 GPU for 120K iterations. We retrain all baseline models with our mixed dataset including MotionX, TED and Champ dataset.  Our HMInterp has similar inference speed and memory cost with AnimateAnyone (50.7s vs. 48.1s, 39.2G vs. 37.5G for 768×768 14-frame videos on A100 40G).

\section{Rule-based Searching}
\label{secsup4}
Our system could support motion retrieval for diverse scenarios by introducing task-specific rules. We implement three types of conditions in this paper: Music2Dance, Action2Motion, and Speech2Gesture. These rules-based searches demonstrate the system's ability to align generated motions with the different conditions. We define the path cost functions as task-specific CLIP-like feature distance, \textit{e.g.}, CLIP-like joint embedding in MotionClip for text\cite{tevet2022motionclip}, ChoreoMaster \cite{chen2021choreomaster} for music, TANGO for speech audio \cite{liu2024tango}. Once the graph and path costs are defined, path searching can be performed using dynamic programming (DP) for offline processing or with efficient Beam Search~\cite{steinbiss1994improvements} for real-time applications. More details on task definitions and task-specific cost designs are provided in the following paragraphs. In addition, we introduce node-level path searching using shortest path algorithms on weighted graphs. Given the target sequence with \( K \) keyframes, we separately search the discontinuous region with the Dijkstra algorithm ~\cite{wang2011application}, using a length scale factor \( D \), which allows the target clip to be \( (1-D) \) times the length of the target, and then re-interpolate the searched path to the target length. 

\paragraph{Music2Dance}
For Music2Dance retrieval, we focus on beat matching to synchronize dance motions with the rhythm of the music. Inspired by AIST++~\cite{li2020learning}, we detect beat points in dance motion by computing local minima in motion velocity and extracting beat information from music. Besides, in the reference video, both motion and music BEAT points are evenly distributed. In addition to beat score, similar to Choreomaster \citep{chen2021choreomaster}, we introduce a structural penalty term. This term penalizes repeated motion patterns excessively. We also adopt the CLIP-like feature distance trained from joint embedding in Choreomaster for music-motion content matching. 

\paragraph{Action2Motion}
For Action2Motion retrieval, we employ a combination of keyword matching and action segmentation to retrieve motions effectively for specific actions. Videos containing multiple actions are first segmented into smaller segments based on action segmentation. Each segment is then tagged with either unsupervised labels (e.g., Action A, B, C) or manually assigned labels (e.g., sitting, walking). During retrieval, each motion segment is assigned two labels: a global action tag and a local ordering tag. The system first matches the global action type and then selects frames from the closest matching local order, ensuring temporal coherence across action segments. For Text2motion, we adopt the CLIP-like feature distance from MotionCLIP. 

\paragraph{Speech2Gesture}
For Speech2Gesture retrieval, we adopt a latent-space-based approach inspired by TANGO \citep{liu2024tango}. We use the pretrained weights from TANGO to calculate the audio-motion difference in latent-space features to determine the optimal path. We minimize the global audio-motion distance using Dynamic Programming (DP). After retrieval and sampling, we adopt a lip-sync model \cite{prajwal2020lip} to post-processing the output. This alignment not only improves visual coherence but also enhances the emotional expressiveness of the generated gestures.

\section{Baseline Settings}
\label{secsup3}
For comparison with previous fully generative human motion video systems, we select state-of-the-art generation methods for various sub-tasks. Specifically, we compare against DanceAnyBEAT \cite{wang2024dance} for dance generation, S2G-Diffusion \cite{he2024co} for gesture generation, and Text2Performer \cite{jiang2023text2performer} for action-to-motion generation. DanceAnyBEAT is a video diffusion model that incorporates audio features via cross-attention and integrates text features into the UNet architecture. S2G-Diffusion is an end-to-end diffusion model designed to generate co-speech gesture videos directly from speech input. Text2Performer generates human motion videos based on action descriptions. For the evaluation, we compare the demo videos available in their repositories with our results through a user study.

\section{Evaluation Metrics}
\label{secsup2}

We use both pixel-level and feature-level evaluations to evaluate the quality of the generated videos. The metrics are Peak Signal-to-Noise Ratio (PSNR), MOtion-based Video Integrity Evaluation (MOVIE) \citep{seshadrinathan2009motion}, Learned Perceptual Image Patch Similarity (LPIPS) \citep{zhang2018unreasonable}, and Fréchet Video Distance (FVD) \citep{skerry2018towards}.

\paragraph{Peak Signal-to-Noise Ratio (PSNR)}
PSNR measures how similar the generated frames are to the ground truth frames at the pixel level. It is based on the mean squared error (MSE) and is expressed in decibels. Higher PSNR values mean less reconstruction error. The formula is:

\begin{equation}
\text{PSNR} = 10 \cdot \log_{10} \left(\frac{{\text{MAX}_{I}^2}}{{\text{MSE}}}\right)
\end{equation}

Here, \(\text{MAX}_{I}\) is the maximum possible pixel value (e.g., 255 for 8-bit images), and MSE is the average squared difference between the original and generated frames.

\paragraph{MOtion-based Video Integrity Evaluation (MOVIE)}
MOVIE evaluates both spatial and temporal differences in video frames to evaluate the video quality. It assesses how well frames are interpolated and how smooth the transitions are. The formula for MOVIE is:

\begin{equation}
\resizebox{0.95\linewidth}{!}{$
\text{MOVIE} = 
\begin{aligned}
& \frac{1}{N} \sum \left[(v_1(t+1) - v_1(t)) - (v_2(t+1) - v_2(t))\right]^2 \\
& + \frac{1}{N} \sum \left(v_1 - v_2\right)^2
\end{aligned}
$}
\end{equation}

Here, \(v_1\) and \(v_2\) represent the ground truth and generated video frames, respectively, and \(N\) is the total number of frames. Lower MOVIE values indicate better video quality.

\paragraph{Learned Perceptual Image Patch Similarity (LPIPS)}
LPIPS measures how similar two images look in terms of features learned by a neural network. It focuses on perceptual quality rather than just pixel accuracy. The formula is:

\begin{equation}
\text{LPIPS}(x, y) = \sum_{l} w_l \left\| \phi_l(x) - \phi_l(y) \right\|_2
\end{equation}

Here, \(x\) and \(y\) are the generated and reference images, \(\phi_l\) is the feature map from the \(l\)-th layer of a pre-trained network, and \(w_l\) is a weight for that layer. Smaller LPIPS scores mean better perceptual similarity. We adopt the pretrained VGG \cite{simonyan2014very} in Pytorch LPIPS as the feature extractor.

\paragraph{Fréchet Video Distance (FVD)}
FVD measures how similar the real and generated videos are in a feature space. It considers both the average features and their variability over time. The formula is:

\begin{equation}
\text{FVD} = || \mu_r - \mu_g ||^2 + \operatorname{Tr}(\Sigma_r + \Sigma_g - 2(\Sigma_r \Sigma_g)^{1/2})
\end{equation}

Here, \(\mu_r\) and \(\mu_g\) are the means of the features for real and generated videos, and \(\Sigma_r\) and \(\Sigma_g\) are their covariances. Lower FVD values indicate better realism and smoother motion in the generated videos. We adopt the pretrained I3D \cite{carreira2017quo} network as the feature extractor.

\section{Ablation study of CLIP and Reference Net}
\label{secsup7}
\begin{table}[h]
\centering
\caption{\textbf{Objective Comparison of CLIP and Reference Net.} Without ReferenceNet and CLIP image encoder, the results perform worse on all objective scores}
\resizebox{0.95\linewidth}{!}{
\begin{tabular}{lcccc}
\textbf{Method}          & \textbf{PSNR $\uparrow$} & \textbf{LPIPS $\downarrow$} & \textbf{MOVIE $\downarrow$} & \textbf{FVD $\downarrow$} \\ \hline
HMInterp ($s=1$)         & 39.53                    & 0.034                       & 39.18                       & 1.210                    \\
w/o CLIP                 & 34.07                    & 0.069                       & 65.47                       & 1.588                    \\
w/o Reference Net        & 32.68                    & 0.110                       & 78.19                       & 2.098                    \\ 
\end{tabular}}
\end{table}

We show that CLIP and Reference Net is necessary to minimize low-level artifacts,  as shown in the Table and Figure, removing CLIP weakens denoising, causing noisier videos, while removing Reference Net results in blurriness.

\begin{figure}[h]
\centering
\includegraphics[width=1.0\columnwidth]{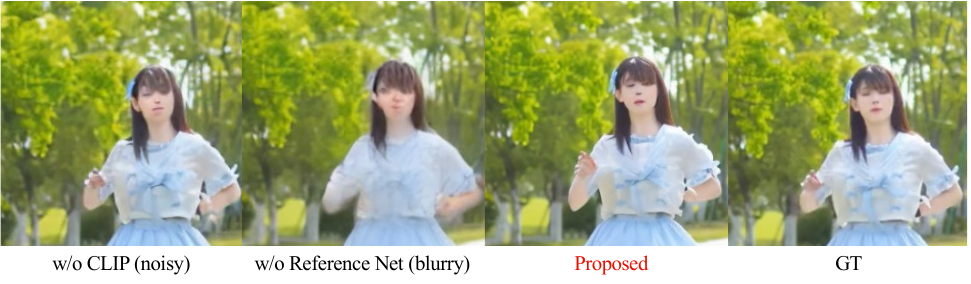}
\caption{\textbf{Subjective Comparison of CLIP and Reference Net.} We remove the CLIP and ReferneceNet on our (denoted as proposed) method. Results show that removing CLIP weakens denosing, and removing Reference Net results in clearly blurriness.}
\label{fig:ablation_clip}
\end{figure}

%% file: main.bbl
\begin{thebibliography}{68}
\providecommand{\natexlab}[1]{#1}
\providecommand{\url}[1]{\texttt{#1}}
\expandafter\ifx\csname urlstyle\endcsname\relax
  \providecommand{\doi}[1]{doi: #1}\else
  \providecommand{\doi}{doi: \begingroup \urlstyle{rm}\Url}\fi

\bibitem[Blattmann et~al.(2023{\natexlab{a}})Blattmann, Dockhorn, Kulal, Mendelevitch, Kilian, Lorenz, Levi, English, Voleti, Letts, et~al.]{blattmann2023stable}
Andreas Blattmann, Tim Dockhorn, Sumith Kulal, Daniel Mendelevitch, Maciej Kilian, Dominik Lorenz, Yam Levi, Zion English, Vikram Voleti, Adam Letts, et~al.
\newblock Stable video diffusion: Scaling latent video diffusion models to large datasets.
\newblock \emph{arXiv preprint arXiv:2311.15127}, 2023{\natexlab{a}}.

\bibitem[Blattmann et~al.(2023{\natexlab{b}})Blattmann, Rombach, Ling, Dockhorn, Kim, Fidler, and Kreis]{blattmann2023align}
Andreas Blattmann, Robin Rombach, Huan Ling, Tim Dockhorn, Seung~Wook Kim, Sanja Fidler, and Karsten Kreis.
\newblock Align your latents: High-resolution video synthesis with latent diffusion models.
\newblock In \emph{Proceedings of the IEEE/CVF Conference on Computer Vision and Pattern Recognition}, pages 22563--22575, 2023{\natexlab{b}}.

\bibitem[Carreira and Zisserman(2017)]{carreira2017quo}
Joao Carreira and Andrew Zisserman.
\newblock Quo vadis, action recognition? a new model and the kinetics dataset.
\newblock In \emph{proceedings of the IEEE Conference on Computer Vision and Pattern Recognition}, pages 6299--6308, 2017.

\bibitem[Chan et~al.(2019)Chan, Ginosar, Zhou, and Efros]{chan2019everybody}
Caroline Chan, Shiry Ginosar, Tinghui Zhou, and Alexei~A Efros.
\newblock Everybody dance now.
\newblock In \emph{Proceedings of the IEEE/CVF international conference on computer vision}, pages 5933--5942, 2019.

\bibitem[Chang et~al.(2023)Chang, Shi, Gao, Xu, Fu, Song, Yan, Zhu, Yang, and Soleymani]{chang2023magicpose}
Di Chang, Yichun Shi, Quankai Gao, Hongyi Xu, Jessica Fu, Guoxian Song, Qing Yan, Yizhe Zhu, Xiao Yang, and Mohammad Soleymani.
\newblock Magicpose: Realistic human poses and facial expressions retargeting with identity-aware diffusion.
\newblock In \emph{Forty-first International Conference on Machine Learning}, 2023.

\bibitem[Chen et~al.(2023)Chen, Xia, He, Zhang, Cun, Yang, Xing, Liu, Chen, Wang, et~al.]{chen2023videocrafter1}
Haoxin Chen, Menghan Xia, Yingqing He, Yong Zhang, Xiaodong Cun, Shaoshu Yang, Jinbo Xing, Yaofang Liu, Qifeng Chen, Xintao Wang, et~al.
\newblock Videocrafter1: Open diffusion models for high-quality video generation.
\newblock \emph{arXiv preprint arXiv:2310.19512}, 2023.

\bibitem[Chen et~al.(2024)Chen, Zhang, Cun, Xia, Wang, Weng, and Shan]{chen2024videocrafter2}
Haoxin Chen, Yong Zhang, Xiaodong Cun, Menghan Xia, Xintao Wang, Chao Weng, and Ying Shan.
\newblock Videocrafter2: Overcoming data limitations for high-quality video diffusion models.
\newblock In \emph{Proceedings of the IEEE/CVF Conference on Computer Vision and Pattern Recognition}, pages 7310--7320, 2024.

\bibitem[Chen et~al.(2021)Chen, Tan, Lei, Zhang, Guo, Zhang, and Hu]{chen2021choreomaster}
Kang Chen, Zhipeng Tan, Jin Lei, Song-Hai Zhang, Yuan-Chen Guo, Weidong Zhang, and Shi-Min Hu.
\newblock Choreomaster: choreography-oriented music-driven dance synthesis.
\newblock \emph{ACM Transactions on Graphics (TOG)}, 40\penalty0 (4):\penalty0 1--13, 2021.

\bibitem[Corona et~al.(2024)Corona, Zanfir, Bazavan, Kolotouros, Alldieck, and Sminchisescu]{corona2024vlogger}
Enric Corona, Andrei Zanfir, Eduard~Gabriel Bazavan, Nikos Kolotouros, Thiemo Alldieck, and Cristian Sminchisescu.
\newblock Vlogger: Multimodal diffusion for embodied avatar synthesis.
\newblock \emph{arXiv preprint arXiv:2403.08764}, 2024.

\bibitem[Danier et~al.(2022)Danier, Zhang, and Bull]{danier2022stmfnet}
Duolikun Danier, Fan Zhang, and David~R. Bull.
\newblock St-mfnet: {A} spatio-temporal multi-flow network for frame interpolation.
\newblock In \emph{Proceedings of the IEEE/CVF Conference on Computer Vision and Pattern Recognition (CVPR)}, pages 3511--3521, 2022.

\bibitem[Danier et~al.(2024)Danier, Zhang, and Bull]{danier2024ldmvfi}
Duolikun Danier, Fan Zhang, and David~R. Bull.
\newblock {LDMVFI:} video frame interpolation with latent diffusion models.
\newblock In \emph{Proceedings of the AAAI Conference on Artificial Intelligence}, pages 1472--1480, 2024.

\bibitem[Esser et~al.(2024)Esser, Kulal, Blattmann, Entezari, M{\"u}ller, Saini, Levi, Lorenz, Sauer, Boesel, et~al.]{esser2024scaling}
Patrick Esser, Sumith Kulal, Andreas Blattmann, Rahim Entezari, Jonas M{\"u}ller, Harry Saini, Yam Levi, Dominik Lorenz, Axel Sauer, Frederic Boesel, et~al.
\newblock Scaling rectified flow transformers for high-resolution image synthesis.
\newblock In \emph{Forty-first International Conference on Machine Learning}, 2024.

\bibitem[Feng et~al.(2025)Feng, Ding, Xia, Niklaus, Abrevaya, Black, and Zhang]{feng2025explorative}
Haiwen Feng, Zheng Ding, Zhihao Xia, Simon Niklaus, Victoria Abrevaya, Michael~J Black, and Xuaner Zhang.
\newblock Explorative inbetweening of time and space.
\newblock In \emph{European Conference on Computer Vision}, pages 378--395. Springer, 2025.

\bibitem[Guo et~al.(2023)Guo, Yang, Rao, Wang, Qiao, Lin, and Dai]{guo2023animatediff}
Yuwei Guo, Ceyuan Yang, Anyi Rao, Yaohui Wang, Yu Qiao, Dahua Lin, and Bo Dai.
\newblock Animatediff: Animate your personalized text-to-image diffusion models without specific tuning.
\newblock \emph{arXiv preprint arXiv:2307.04725}, 2023.

\bibitem[He et~al.(2024{\natexlab{a}})He, Huang, Zhang, Lin, Wu, Yang, Li, Chen, Xu, and Wu]{he2024co}
Xu He, Qiaochu Huang, Zhensong Zhang, Zhiwei Lin, Zhiyong Wu, Sicheng Yang, Minglei Li, Zhiyi Chen, Songcen Xu, and Xiaofei Wu.
\newblock Co-speech gesture video generation via motion-decoupled diffusion model.
\newblock In \emph{Proceedings of the IEEE/CVF Conference on Computer Vision and Pattern Recognition}, pages 2263--2273, 2024{\natexlab{a}}.

\bibitem[He et~al.(2024{\natexlab{b}})He, Huang, Zhang, Lin, Wu, Yang, Li, Chen, Xu, and Wu]{he2024cospeech}
Xu He, Qiaochu Huang, Zhensong Zhang, Zhiwei Lin, Zhiyong Wu, Sicheng Yang, Minglei Li, Zhiyi Chen, Songcen Xu, and Xiaofei Wu.
\newblock Co-speech gesture video generation via motion-decoupled diffusion model.
\newblock \emph{arXiv preprint arXiv:2404.01862}, 2024{\natexlab{b}}.

\bibitem[Hong et~al.(2022)Hong, Ding, Zheng, Liu, and Tang]{hong2022cogvideo}
Wenyi Hong, Ming Ding, Wendi Zheng, Xinghan Liu, and Jie Tang.
\newblock Cogvideo: Large-scale pretraining for text-to-video generation via transformers.
\newblock \emph{arXiv preprint arXiv:2205.15868}, 2022.

\bibitem[Hu et~al.(2023)Hu, Gao, Zhang, Sun, Zhang, and Bo]{hu2023animate}
Li Hu, Xin Gao, Peng Zhang, Ke Sun, Bang Zhang, and Liefeng Bo.
\newblock Animate anyone: Consistent and controllable image-to-video synthesis for character animation.
\newblock \emph{arXiv preprint arXiv:2311.17117}, 2023.

\bibitem[Huang et~al.(2022)Huang, Zhang, Heng, Shi, and Zhou]{huang2022real}
Zhewei Huang, Tianyuan Zhang, Wen Heng, Boxin Shi, and Shuchang Zhou.
\newblock Real-time intermediate flow estimation for video frame interpolation.
\newblock In \emph{European Conference on Computer Vision}, pages 624--642. Springer, 2022.

\bibitem[Huang et~al.(2024)Huang, Tang, Zhang, Cun, Cao, Li, and Lee]{huang2024make}
Ziyao Huang, Fan Tang, Yong Zhang, Xiaodong Cun, Juan Cao, Jintao Li, and Tong-Yee Lee.
\newblock Make-your-anchor: A diffusion-based 2d avatar generation framework.
\newblock In \emph{Proceedings of the IEEE/CVF Conference on Computer Vision and Pattern Recognition}, pages 6997--7006, 2024.

\bibitem[Jain et~al.(2024)Jain, Watson, Tabellion, Poole, Kontkanen, et~al.]{jain2024video}
Siddhant Jain, Daniel Watson, Eric Tabellion, Ben Poole, Janne Kontkanen, et~al.
\newblock Video interpolation with diffusion models.
\newblock In \emph{Proceedings of the IEEE/CVF Conference on Computer Vision and Pattern Recognition}, pages 7341--7351, 2024.

\bibitem[Jiang et~al.(2018)Jiang, Sun, Jampani, Yang, Learned{-}Miller, and Kautz]{jiang2018superslomo}
Huaizu Jiang, Deqing Sun, Varun Jampani, Ming{-}Hsuan Yang, Erik~G. Learned{-}Miller, and Jan Kautz.
\newblock Super slomo: High quality estimation of multiple intermediate frames for video interpolation.
\newblock In \emph{Proceedings of the IEEE/CVF Conference on Computer Vision and Pattern Recognition (CVPR)}, pages 9000--9008, 2018.

\bibitem[Jiang et~al.(2023)Jiang, Yang, Koh, Wu, Loy, and Liu]{jiang2023text2performer}
Yuming Jiang, Shuai Yang, Tong~Liang Koh, Wayne Wu, Chen~Change Loy, and Ziwei Liu.
\newblock Text2performer: Text-driven human video generation.
\newblock In \emph{Proceedings of the IEEE/CVF International Conference on Computer Vision}, pages 22747--22757, 2023.

\bibitem[Kamel et~al.(2008)Kamel, Ebrahimnezhad, and Ebrahimi]{kamel2008moving}
Saeed Kamel, Hossein Ebrahimnezhad, and Afshin Ebrahimi.
\newblock Moving object removal in video sequence and background restoration using kalman filter.
\newblock In \emph{2008 International Symposium on Telecommunications}, pages 580--585. IEEE, 2008.

\bibitem[Kong et~al.(2022)Kong, Jiang, Luo, Chu, Huang, Tai, Wang, and Yang]{kong2022ifrnet}
Lingtong Kong, Boyuan Jiang, Donghao Luo, Wenqing Chu, Xiaoming Huang, Ying Tai, Chengjie Wang, and Jie Yang.
\newblock Ifrnet: Intermediate feature refine network for efficient frame interpolation.
\newblock In \emph{Proceedings of the IEEE/CVF Conference on Computer Vision and Pattern Recognition}, pages 1969--1978, 2022.

\bibitem[Kovar et~al.(2008)Kovar, Gleicher, and Pighin]{kovar2008motiongraphs}
Lucas Kovar, Michael Gleicher, and Fr{\'{e}}d{\'{e}}ric~H. Pighin.
\newblock Motion graphs.
\newblock pages 51:1--51:10, 2008.

\bibitem[Li et~al.(2020)Li, Yin, Chu, Zhou, Wang, Fidler, and Li]{li2020learning}
Jiaman Li, Yihang Yin, Hang Chu, Yi Zhou, Tingwu Wang, Sanja Fidler, and Hao Li.
\newblock Learning to generate diverse dance motions with transformer.
\newblock \emph{arXiv preprint arXiv:2008.08171}, 2020.

\bibitem[Li et~al.(2023)Li, Zhu, Han, Hou, Guo, and Cheng]{li2023amt}
Zhen Li, Zuo{-}Liang Zhu, Linghao Han, Qibin Hou, Chun{-}Le Guo, and Ming{-}Ming Cheng.
\newblock {AMT:} all-pairs multi-field transforms for efficient frame interpolation.
\newblock In \emph{Proceedings of the IEEE/CVF Conference on Computer Vision and Pattern Recognition (CVPR)}, pages 9801--9810, 2023.

\bibitem[Lin et~al.(2024{\natexlab{a}})Lin, Jiang, Liang, Zhong, Yang, and Zheng]{lin2024cyberhost}
Gaojie Lin, Jianwen Jiang, Chao Liang, Tianyun Zhong, Jiaqi Yang, and Yanbo Zheng.
\newblock Cyberhost: Taming audio-driven avatar diffusion model with region codebook attention.
\newblock \emph{arXiv preprint arXiv:2409.01876}, 2024{\natexlab{a}}.

\bibitem[Lin et~al.(2024{\natexlab{b}})Lin, Zeng, Lu, Cai, Zhang, Wang, and Zhang]{lin2024motion}
Jing Lin, Ailing Zeng, Shunlin Lu, Yuanhao Cai, Ruimao Zhang, Haoqian Wang, and Lei Zhang.
\newblock Motion-x: A large-scale 3d expressive whole-body human motion dataset.
\newblock \emph{Advances in Neural Information Processing Systems}, 36, 2024{\natexlab{b}}.

\bibitem[Liu et~al.(2022{\natexlab{a}})Liu, Iwamoto, Zhu, Li, Zhou, Bozkurt, and Zheng]{liu2022disco}
Haiyang Liu, Naoya Iwamoto, Zihao Zhu, Zhengqing Li, You Zhou, Elif Bozkurt, and Bo Zheng.
\newblock Disco: Disentangled implicit content and rhythm learning for diverse co-speech gestures synthesis.
\newblock In \emph{Proceedings of the 30th ACM International Conference on Multimedia}, pages 3764--3773, 2022{\natexlab{a}}.

\bibitem[Liu et~al.(2022{\natexlab{b}})Liu, Zhu, Iwamoto, Peng, Li, Zhou, Bozkurt, and Zheng]{liu2022beat}
Haiyang Liu, Zihao Zhu, Naoya Iwamoto, Yichen Peng, Zhengqing Li, You Zhou, Elif Bozkurt, and Bo Zheng.
\newblock Beat: A large-scale semantic and emotional multi-modal dataset for conversational gestures synthesis.
\newblock \emph{arXiv preprint arXiv:2203.05297}, 2022{\natexlab{b}}.

\bibitem[Liu et~al.(2024{\natexlab{a}})Liu, Yang, Akiyama, Huang, Li, Kuriyama, and Taketomi]{liu2024tango}
Haiyang Liu, Xingchao Yang, Tomoya Akiyama, Yuantian Huang, Qiaoge Li, Shigeru Kuriyama, and Takafumi Taketomi.
\newblock Tango: Co-speech gesture video reenactment with hierarchical audio motion embedding and diffusion interpolation.
\newblock \emph{arXiv preprint arXiv:2410.04221}, 2024{\natexlab{a}}.

\bibitem[Liu et~al.(2024{\natexlab{b}})Liu, Zhu, Becherini, Peng, Su, Zhou, Zhe, Iwamoto, Zheng, and Black]{liu2024emage}
Haiyang Liu, Zihao Zhu, Giorgio Becherini, Yichen Peng, Mingyang Su, You Zhou, Xuefei Zhe, Naoya Iwamoto, Bo Zheng, and Michael~J Black.
\newblock Emage: Towards unified holistic co-speech gesture generation via expressive masked audio gesture modeling.
\newblock In \emph{Proceedings of the IEEE/CVF Conference on Computer Vision and Pattern Recognition}, pages 1144--1154, 2024{\natexlab{b}}.

\bibitem[Liu et~al.(2017)Liu, Yeh, Tang, Liu, and Agarwala]{liu2017voxelflow}
Ziwei Liu, Raymond~A. Yeh, Xiaoou Tang, Yiming Liu, and Aseem Agarwala.
\newblock Video frame synthesis using deep voxel flow.
\newblock In \emph{Proceedings of the IEEE/CVF International Conference on Computer Vision (ICCV)}, pages 4473--4481, 2017.

\bibitem[Lu et~al.(2022)Lu, Wu, Lin, Lu, and Jia]{lu2022video}
Liying Lu, Ruizheng Wu, Huaijia Lin, Jiangbo Lu, and Jiaya Jia.
\newblock Video frame interpolation with transformer.
\newblock In \emph{Proceedings of the IEEE/CVF Conference on Computer Vision and Pattern Recognition}, pages 3532--3542, 2022.

\bibitem[Niklaus and Liu(2018)]{niklaus2018contextinterp}
Simon Niklaus and Feng Liu.
\newblock Context-aware synthesis for video frame interpolation.
\newblock In \emph{Proceedings of the IEEE/CVF Conference on Computer Vision and Pattern Recognition (CVPR)}, pages 1701--1710, 2018.

\bibitem[Niklaus and Liu(2020)]{niklaus2020softmaxinterp}
Simon Niklaus and Feng Liu.
\newblock Softmax splatting for video frame interpolation.
\newblock In \emph{Proceedings of the IEEE/CVF Conference on Computer Vision and Pattern Recognition (CVPR)}, pages 5436--5445, 2020.

\bibitem[Pan et~al.(2024)Pan, Xu, Huang, Singh, Zhou, Guibas, and Yang]{pan2024actanywhere}
Boxiao Pan, Zhan Xu, Chun-Hao~Paul Huang, Krishna~Kumar Singh, Yang Zhou, Leonidas~J Guibas, and Jimei Yang.
\newblock Actanywhere: Subject-aware video background generation.
\newblock \emph{arXiv preprint arXiv:2401.10822}, 2024.

\bibitem[Park et~al.(2020)Park, Ko, Lee, and Kim]{park2020bmbc}
Junheum Park, Keunsoo Ko, Chul Lee, and Chang{-}Su Kim.
\newblock {BMBC:} bilateral motion estimation with bilateral cost volume for video interpolation.
\newblock In \emph{European Conference on Computer Vision}, pages 109--125, 2020.

\bibitem[Park et~al.(2021)Park, Lee, and Kim]{park2021asymmetricinterp}
Junheum Park, Chul Lee, and Chang{-}Su Kim.
\newblock Asymmetric bilateral motion estimation for video frame interpolation.
\newblock In \emph{Proceedings of the IEEE/CVF International Conference on Computer Vision (ICCV)}, pages 14519--14528, 2021.

\bibitem[Prajwal et~al.(2020)Prajwal, Mukhopadhyay, Namboodiri, and Jawahar]{prajwal2020lip}
KR Prajwal, Rudrabha Mukhopadhyay, Vinay~P Namboodiri, and CV Jawahar.
\newblock A lip sync expert is all you need for speech to lip generation in the wild.
\newblock In \emph{Proceedings of the 28th ACM international conference on multimedia}, pages 484--492, 2020.

\bibitem[Radford et~al.(2021)Radford, Kim, Hallacy, Ramesh, Goh, Agarwal, Sastry, Askell, Mishkin, Clark, et~al.]{radford2021learning}
Alec Radford, Jong~Wook Kim, Chris Hallacy, Aditya Ramesh, Gabriel Goh, Sandhini Agarwal, Girish Sastry, Amanda Askell, Pamela Mishkin, Jack Clark, et~al.
\newblock Learning transferable visual models from natural language supervision.
\newblock In \emph{International conference on machine learning}, pages 8748--8763. PMLR, 2021.

\bibitem[Ramesh et~al.(2021)Ramesh, Pavlov, Goh, Gray, Voss, Radford, Chen, and Sutskever]{ramesh2021zero}
Aditya Ramesh, Mikhail Pavlov, Gabriel Goh, Scott Gray, Chelsea Voss, Alec Radford, Mark Chen, and Ilya Sutskever.
\newblock Zero-shot text-to-image generation.
\newblock In \emph{International conference on machine learning}, pages 8821--8831. Pmlr, 2021.

\bibitem[Reda et~al.(2022)Reda, Kontkanen, Tabellion, Sun, Pantofaru, and Curless]{reda2022film}
Fitsum Reda, Janne Kontkanen, Eric Tabellion, Deqing Sun, Caroline Pantofaru, and Brian Curless.
\newblock {FILM:} frame interpolation for large motion.
\newblock In \emph{European Conference on Computer Vision}, pages 250--266. Springer, 2022.

\bibitem[Rombach et~al.(2022)Rombach, Blattmann, Lorenz, Esser, and Ommer]{rombach2022high}
Robin Rombach, Andreas Blattmann, Dominik Lorenz, Patrick Esser, and Bj{\"o}rn Ommer.
\newblock High-resolution image synthesis with latent diffusion models.
\newblock In \emph{Proceedings of the IEEE/CVF conference on computer vision and pattern recognition}, pages 10684--10695, 2022.

\bibitem[Seshadrinathan and Bovik(2009)]{seshadrinathan2009motion}
Kalpana Seshadrinathan and Alan~Conrad Bovik.
\newblock Motion tuned spatio-temporal quality assessment of natural videos.
\newblock \emph{IEEE transactions on image processing}, 19\penalty0 (2):\penalty0 335--350, 2009.

\bibitem[Shafir et~al.(2023)Shafir, Tevet, Kapon, and Bermano]{shafir2023human}
Yonatan Shafir, Guy Tevet, Roy Kapon, and Amit~H Bermano.
\newblock Human motion diffusion as a generative prior.
\newblock \emph{arXiv preprint arXiv:2303.01418}, 2023.

\bibitem[Sim et~al.(2021)Sim, Oh, and Kim]{sim2021xvfi}
Hyeonjun Sim, Jihyong Oh, and Munchurl Kim.
\newblock {XVFI:} extreme video frame interpolation.
\newblock In \emph{Proceedings of the IEEE/CVF International Conference on Computer Vision (ICCV)}, pages 14469--14478, 2021.

\bibitem[Simonyan and Zisserman(2014)]{simonyan2014very}
Karen Simonyan and Andrew Zisserman.
\newblock Very deep convolutional networks for large-scale image recognition.
\newblock \emph{arXiv preprint arXiv:1409.1556}, 2014.

\bibitem[Skerry-Ryan et~al.(2018)Skerry-Ryan, Battenberg, Xiao, Wang, Stanton, Shor, Weiss, Clark, and Saurous]{skerry2018towards}
RJ Skerry-Ryan, Eric Battenberg, Ying Xiao, Yuxuan Wang, Daisy Stanton, Joel Shor, Ron Weiss, Rob Clark, and Rif~A Saurous.
\newblock Towards end-to-end prosody transfer for expressive speech synthesis with tacotron.
\newblock In \emph{international conference on machine learning}, pages 4693--4702. PMLR, 2018.

\bibitem[Steinbiss et~al.(1994)Steinbiss, Tran, and Ney]{steinbiss1994improvements}
Volker Steinbiss, Bach-Hiep Tran, and Hermann Ney.
\newblock Improvements in beam search.
\newblock In \emph{ICSLP}, pages 2143--2146, 1994.

\bibitem[Tevet et~al.(2022)Tevet, Gordon, Hertz, Bermano, and Cohen-Or]{tevet2022motionclip}
Guy Tevet, Brian Gordon, Amir Hertz, Amit~H Bermano, and Daniel Cohen-Or.
\newblock Motionclip: Exposing human motion generation to clip space.
\newblock In \emph{European Conference on Computer Vision}, pages 358--374. Springer, 2022.

\bibitem[Wang et~al.(2011)Wang, Yu, and Yuan]{wang2011application}
Huijuan Wang, Yuan Yu, and Quanbo Yuan.
\newblock Application of dijkstra algorithm in robot path-planning.
\newblock In \emph{2011 second international conference on mechanic automation and control engineering}, pages 1067--1069. IEEE, 2011.

\bibitem[Wang et~al.(2024{\natexlab{a}})Wang, Wang, Liu, and Cai]{wang2024dance}
Xuanchen Wang, Heng Wang, Dongnan Liu, and Weidong Cai.
\newblock Dance any beat: Blending beats with visuals in dance video generation.
\newblock \emph{arXiv preprint arXiv:2405.09266}, 2024{\natexlab{a}}.

\bibitem[Wang et~al.(2024{\natexlab{b}})Wang, Zhang, Gao, Wang, Zhou, Zhang, Yan, and Sang]{wang2024unianimate}
Xiang Wang, Shiwei Zhang, Changxin Gao, Jiayu Wang, Xiaoqiang Zhou, Yingya Zhang, Luxin Yan, and Nong Sang.
\newblock Unianimate: Taming unified video diffusion models for consistent human image animation.
\newblock \emph{arXiv preprint arXiv:2406.01188}, 2024{\natexlab{b}}.

\bibitem[Wu et~al.(2022)Wu, Wen, and Chen]{wu2022optimizing}
Yue Wu, Qiang Wen, and Qifeng Chen.
\newblock Optimizing video prediction via video frame interpolation.
\newblock In \emph{Proceedings of the IEEE/CVF Conference on Computer Vision and Pattern Recognition}, pages 17814--17823, 2022.

\bibitem[Xing et~al.(2024)Xing, Liu, Xia, Zhang, Wang, Shan, and Wong]{xing2024tooncrafter}
Jinbo Xing, Hanyuan Liu, Menghan Xia, Yong Zhang, Xintao Wang, Ying Shan, and Tien-Tsin Wong.
\newblock Tooncrafter: Generative cartoon interpolation.
\newblock \emph{arXiv preprint arXiv:2405.17933}, 2024.

\bibitem[Xing et~al.(2025)Xing, Xia, Zhang, Chen, Yu, Liu, Liu, Wang, Shan, and Wong]{xing2025dynamicrafter}
Jinbo Xing, Menghan Xia, Yong Zhang, Haoxin Chen, Wangbo Yu, Hanyuan Liu, Gongye Liu, Xintao Wang, Ying Shan, and Tien-Tsin Wong.
\newblock Dynamicrafter: Animating open-domain images with video diffusion priors.
\newblock In \emph{European Conference on Computer Vision}, pages 399--417. Springer, 2025.

\bibitem[Xu et~al.(2024)Xu, Zhang, Liew, Yan, Liu, Zhang, Feng, and Shou]{xu2024magicanimate}
Zhongcong Xu, Jianfeng Zhang, Jun~Hao Liew, Hanshu Yan, Jia-Wei Liu, Chenxu Zhang, Jiashi Feng, and Mike~Zheng Shou.
\newblock Magicanimate: Temporally consistent human image animation using diffusion model.
\newblock In \emph{Proceedings of the IEEE/CVF Conference on Computer Vision and Pattern Recognition}, pages 1481--1490, 2024.

\bibitem[Xue et~al.(2019)Xue, Chen, Wu, Wei, and Freeman]{xue2019vetf}
Tianfan Xue, Baian Chen, Jiajun Wu, Donglai Wei, and William~T. Freeman.
\newblock Video enhancement with task-oriented flow.
\newblock \emph{International Journal of Computer Vision}, 127\penalty0 (8):\penalty0 1106--1125, 2019.

\bibitem[Yang et~al.(2024)Yang, Teng, Zheng, Ding, Huang, Xu, Yang, Hong, Zhang, Feng, et~al.]{yang2024cogvideox}
Zhuoyi Yang, Jiayan Teng, Wendi Zheng, Ming Ding, Shiyu Huang, Jiazheng Xu, Yuanming Yang, Wenyi Hong, Xiaohan Zhang, Guanyu Feng, et~al.
\newblock Cogvideox: Text-to-video diffusion models with an expert transformer.
\newblock \emph{arXiv preprint arXiv:2408.06072}, 2024.

\bibitem[Yi et~al.(2023)Yi, Liang, Liu, Cao, Wen, Bolkart, Tao, and Black]{talkshow:yi2022generating}
Hongwei Yi, Hualin Liang, Yifei Liu, Qiong Cao, Yandong Wen, Timo Bolkart, Dacheng Tao, and Michael~J Black.
\newblock Generating holistic 3d human motion from speech.
\newblock In \emph{CVPR}, 2023.

\bibitem[Zhang et~al.(2023)Zhang, Rao, and Agrawala]{zhang2023adding}
Lvmin Zhang, Anyi Rao, and Maneesh Agrawala.
\newblock Adding conditional control to text-to-image diffusion models.
\newblock In \emph{Proceedings of the IEEE/CVF International Conference on Computer Vision}, pages 3836--3847, 2023.

\bibitem[Zhang et~al.(2018)Zhang, Isola, Efros, Shechtman, and Wang]{zhang2018unreasonable}
Richard Zhang, Phillip Isola, Alexei~A Efros, Eli Shechtman, and Oliver Wang.
\newblock The unreasonable effectiveness of deep features as a perceptual metric.
\newblock In \emph{Proceedings of the IEEE conference on computer vision and pattern recognition}, pages 586--595, 2018.

\bibitem[Zhang et~al.(2024)Zhang, Gu, Wang, Wang, Cheng, Zhu, and Zou]{zhang2024mimicmotion}
Yuang Zhang, Jiaxi Gu, Li-Wen Wang, Han Wang, Junqi Cheng, Yuefeng Zhu, and Fangyuan Zou.
\newblock Mimicmotion: High-quality human motion video generation with confidence-aware pose guidance.
\newblock \emph{arXiv preprint arXiv:2406.19680}, 2024.

\bibitem[Zhou et~al.(2022)Zhou, Yang, Li, Saito, Aneja, and Kalogerakis]{zhou2022audio}
Yang Zhou, Jimei Yang, Dingzeyu Li, Jun Saito, Deepali Aneja, and Evangelos Kalogerakis.
\newblock Audio-driven neural gesture reenactment with video motion graphs.
\newblock In \emph{Proceedings of the IEEE/CVF Conference on Computer Vision and Pattern Recognition}, pages 3418--3428, 2022.

\bibitem[Zhu et~al.(2024)Zhu, Chen, Dai, Su, Xu, Cao, Yao, Zhu, and Zhu]{zhu2024champ}
Shenhao Zhu, Junming~Leo Chen, Zuozhuo Dai, Qingkun Su, Yinghui Xu, Xun Cao, Yao Yao, Hao Zhu, and Siyu Zhu.
\newblock Champ: Controllable and consistent human image animation with 3d parametric guidance.
\newblock \emph{arXiv preprint arXiv:2403.14781}, 2024.

\end{thebibliography}
